\documentclass[acmsmall]{acmart}
\AtBeginDocument{%
  \providecommand\BibTeX{{%
    \normalfont B\kern-0.5em{\scshape i\kern-0.25em b}\kern-0.8em\TeX}}}

\setcopyright{acmcopyright}
\copyrightyear{2018}
\acmYear{2018}
\acmDOI{XXXXXXX.XXXXXXX}

\acmJournal{JACM}
\acmVolume{37}
\acmNumber{4}
\acmArticle{111}
\acmMonth{8}

\usepackage{natbib}
\usepackage[utf8]{inputenc} 
\usepackage[T1]{fontenc}
\usepackage{amsthm}
\usepackage{tikz}
\usepackage[edges]{forest}
\usepackage{hyperref}
\usepackage{graphicx}
\usepackage{wrapfig}

\definecolor{hiddendraw}{RGB}{205, 44, 36}
\definecolor{hidden-blue}{RGB}{194,232,247}
\definecolor{hidden-orange}{RGB}{243,202,120}
\definecolor{hidden-yellow}{RGB}{242,244,193}
\theoremstyle{definition}

\begin{document}


\title{DeepSeek-Inspired Exploration of RL-based LLMs and Synergy with Wireless Networks: A Survey}

\author{Yu Qiao}
\email{qiaoyu@khu.ac.kr}
\orcid{0003-4045-8473}
\affiliation{%
  \institution{Kyung Hee University}
  \city{Yongin-si 17104}
  \country{Republic of Korea}
}

\author{Phuong-Nam Tran}
\email{tpnam0901@khu.ac.kr}
\affiliation{%
  \institution{Kyung Hee University}
  \city{Yongin-si 17104}
  \country{Republic of Korea}}

\author{Ji Su Yoon}
\email{yjs9512@khu.ac.kr}
\affiliation{%
  \institution{Kyung Hee University}
  \city{Yongin-si 17104}
  \country{Republic of Korea}
}

\author{Loc X. Nguyen}
\email{xuanloc088@khu.ac.kr}
\affiliation{%
 \institution{Kyung Hee University}
 \city{Yongin-si 17104}
 \country{Republic of Korea}}

\author{Eui-Nam Huh}
\email{johnhuh@khu.ac.kr}
\affiliation{%
\institution{Kyung Hee University}
\city{Yongin-si 17104}
\country{Republic of Korea}}

\author{Dusit Niyato}
\email{dniyato@ntu.edu.sg}
\affiliation{%
  \institution{Nanyang Technological University}
  \country{Singapore}}

\author{Choong Seon Hong}
\email{cshong@khu.ac.kr}
\affiliation{%
  \institution{Kyung Hee University}
  \city{Yongin-si 17104}
  \country{Republic of Korea}}

\renewcommand{\shortauthors}{Y. Qiao et al.}

\begin{abstract}
    Reinforcement learning (RL)-based large language models (LLMs), such as ChatGPT, DeepSeek, and Grok-3, have attracted widespread attention for their remarkable capabilities in multimodal data understanding. Meanwhile, the rapid expansion of information services has led to a growing demand for AI-enabled wireless networks. The open-source DeepSeek models are famous for their innovative designs, such as large-scale pure RL and cost-efficient training, which make them well-suited for practical deployment in wireless networks. By integrating DeepSeek-style LLMs with wireless infrastructures, a synergistic opportunity arises: the DeepSeek-style LLMs enhance network optimization with strong reasoning and decision-making abilities, while wireless infrastructure enables the broad deployment of these models. Motivated by this convergence, this survey presents a comprehensive DeepSeek-inspired exploration of RL-based LLMs in the context of wireless networks. We begin by reviewing key techniques behind network optimization to establish a foundation for understanding DeepSeek-style LLM integration. Next, we examine recent advancements in RL-based LLMs, using DeepSeek models as a representative example. Building on this, we explore the synergy between the two domains, highlighting motivations, challenges, and potential solutions. Finally, we highlight emerging directions for integrating LLMs with wireless networks, such as quantum, on-device, and neural-symbolic LLM models, as well as embodied AI agents. Overall, this survey offers a comprehensive examination of the interplay between DeepSeek-style LLMs and wireless networks, demonstrating how these domains can mutually enhance each other to drive innovation.
\end{abstract}





\maketitle

\section{Introduction}
Wireless communications~\cite{liu2024survey,loc2024semantic_communication} have become the backbone of modern society, playing an essential role in a wide range of applications, including edge networks~\cite{qiao2025towards,xu2023edge,qiao2024federated}, satellite communication~\cite{talgat2024stochastic,fourati2021artificial,hassan2024semantic}, Internet of Things (IoT)~\cite{qiao2023mp,zhang2015secure}, drone communication~\cite{mozaffari2018communications,hayajneh2016optimal}, and vehicular networks (V2X)~\cite{gerla2011vehicular,nguyen2024efficient}. However, as the demand for efficient information services continues to surge, the pressure on wireless networks has increased, prompting numerous efforts to develop advanced algorithms aimed at alleviating the network burden~\cite{jin2020novel,cichon2016energy,ge2024ris,zou2024energy}. Recently, researchers have increasingly leveraged artificial intelligence (AI) technologies to address key challenges in wireless networks, including efficient resource management~\cite{adhikary2023integrated,khoramnejad2025generative,raha2023generative}, intelligent network architecture~\cite{adhikary2024holographic,cao2017aif,adhikary2023intelligent}, and the design of seamless connectivity and quality service~\cite{raha2023segment,nguyen2024optimizing}. These advancements underscore AI's pivotal role in enhancing the efficiency and intelligence of wireless networks.


More recently, significant progress has been made in the field of natural language processing (NLP), with large language models (LLMs) such as ChatGPT~\cite{achiam2023gpt}, Gemini~\cite{team2023gemini}, and DeepSeek~\cite{guo_daya2025deepseekr1} gaining substantial attention for their advanced language understanding and human-like text generation capabilities. The emergence of these foundational models marks an important milestone in the development of AI, providing highly effective solutions for various complex NLP tasks. For instance, LLMs have been utilized for autonomous code generation~\cite{dong2024self}, medical diagnosis~\cite{garg2023exploring}, and intelligent tutoring systems~\cite{vujinovic2024using}, demonstrating their versatility across diverse domains. Notably, most state-of-the-art LLMs currently integrate reinforcement learning (RL)~\cite{milani2024explainable} to further enhance their performance by leveraging human feedback during the post-training phase~\cite{wang2024reinforcement}. Notable examples include InstructGPT~\cite{ouyang_long2022RLHF}, GPT-4~\cite{achiam2023gpt}, Starling-7B~\cite{zhu2024starling}, and ChatGLM~\cite{glm2024chatglm}, all of which incorporate RL to align outputs with user preferences better and reduce hallucinations. Among them, DeepSeek stands out as a pioneering effort demonstrating that purely RL-based training, even without supervised data, can significantly improve LLM reasoning capabilities~\cite{guo_daya2025deepseekr1}. Moreover, its open-source release lowers the barrier to entry for researchers and practitioners, offering greater accessibility, transparency, and customizability. Motivated by these unique features, this paper adopts DeepSeek as a representative and inspirational case study to explore the broader landscape of RL-based LLMs.

Beyond NLP, groundbreaking advancements have also emerged in computer vision. Meta AI’s segment anything model (SAM)~\cite{kirillov2023segment} and its successor, SAM-2~\cite{ravi2024sam}, have demonstrated remarkable progress in both image and video processing. These developments underscore the growing convergence of vision and language models, paving the way for more advanced multimodal AI systems. With their strong generalization capabilities and adaptability, these foundational models are being explored for transformative applications across various interdisciplinary fields. In the context of wireless networks, leveraging the advantages of RL in long-term optimization, autonomous exploration, and scalability~\cite{lei2020deep,ladosz2022exploration,ei2024deep}, RL-based LLMs offer a promising path for optimizing wireless networks. They can enable intelligent decision-making, efficient resource allocation, and autonomous adaptation to dynamic environments~\cite{chen2024adaptive}. For instance, \cite{sevim2024large} incorporates RL into the training of an LLM-assisted agent, enabling it to determine the optimal placement and orientation of a base station in urban environments to maximize coverage. In addition, \cite{chen2024adaptive} leverages the RL framework to optimize LLM deployment in edge computing environments, effectively balancing inference performance and computational load by determining the optimal split point. 


Despite these promising prospects, integrating RL-LLMs into wireless networks still presents several challenges. First, the dynamic and stochastic nature of wireless environments requires models to effectively balance exploration and exploitation while ensuring real-time adaptability~\cite{bai2023towards,wang2024load}. However, LLMs are well known for being large-scale and resource-intensive, requiring substantial computational resources even during inference. For example, training models such as LLaMA (65 billion parameters) takes 2,048 NVIDIA A100 GPUs running for 21 days~\cite{touvron2023llama}, while training GPT-3 (175 billion parameters) takes 1,024 NVIDIA V100 GPUs for 34 days~\cite{zhao2024optimizing}. Moreover, inference remains computationally expensive; for instance, when running inference on 128 A100 GPUs, the cost of processing an 8k sequence with GPT-4 is approximately \$0.0049 per 1,000 tokens~\cite{Wallstreetcn2024}. Second, optimizing large-scale wireless networks requires efficient resource management, including bandwidth, computational power, energy consumption, and spectrum allocations~\cite{chen2024survey,cao2017towards,wang2024load}. However, the computational demands of LLMs are typically associated with high energy consumption, making them challenging to deploy on energy-constrained wireless devices, such as edge devices or IoT nodes. Third, due to the dynamic nature of wireless environments, wireless devices need to adapt to fluctuating network conditions, varying traffic loads, and interference~\cite{letaief2006dynamic,ji2015wireless}. However, since LLMs are typically pre-trained and difficult to modify, their application in wireless environments requires careful consideration of their ability to adapt to such changes, including changes in channel states and network conditions. Finally, deep learning models are widely known to be vulnerable to various attacks, such as adversarial~\cite{qiao2024noms_fedalc,goodfellow2014explaining}, backdoor~\cite{liu2020reflection,gao2020backdoor,doan2021lira}, and Byzantine attacks~\cite{cui2024resilient,li2024blades,wang2024rfvir}, which can mislead the learning process or manipulate model outputs. In RL-based LLMs, the iterative interactions with the environment or humans make them particularly vulnerable to harmful feedback or adversaries. Therefore, ensuring the robustness and security of RL-LLMs against attacks is also a crucial concern for maintaining network reliability. 

\begin{figure*}[t]
    \centering
    \includegraphics[width=0.93\linewidth]{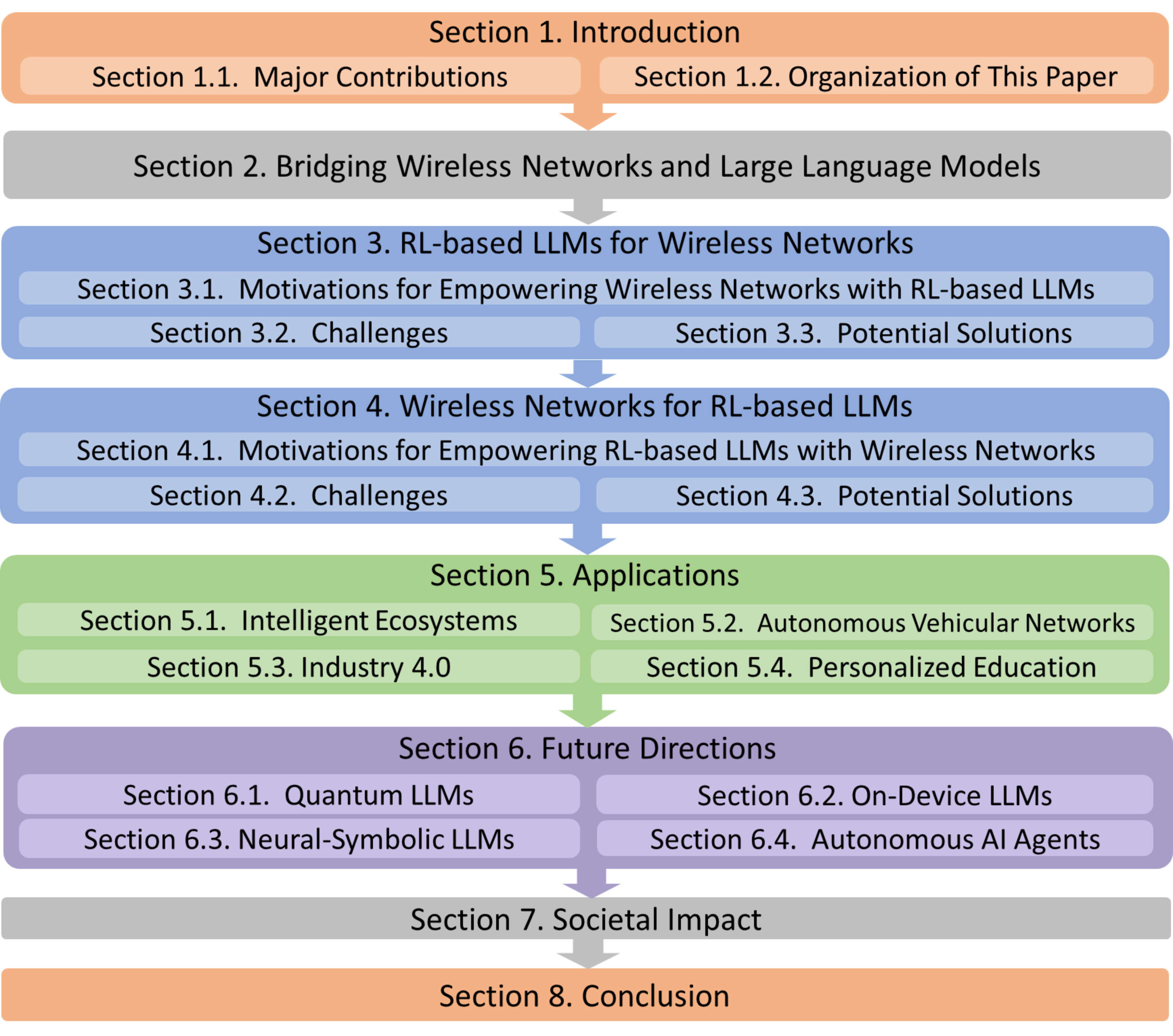}
    \caption{Organization of the Survey.}
    \label{fig:paper_structure}
    \vspace{-7px}
\end{figure*}

On the other hand, wireless networks can also empower LLMs by serving as a robust infrastructure for efficient training, deployment, and inference, particularly in decentralized and edge computing environments. By leveraging the advantages of wireless networks, LLMs can achieve real-time data acquisition, low-latency interactions, large-scale distributed training, and scalable model adaptation in dynamic scenarios. In particular, RL-based strategies for training LLMs are well-suited to such environments, as their effectiveness relies on continuous interaction with the environment and the ability to adapt in response to real-time feedback. For instance, edge computing~\cite{shi2016edge,qiao2023mp,qiao2023framework} and federated learning~\cite{mcmahan2017communication,qiao2024federated,qiao2024logit,thuy2024fl} frameworks can bring RL-LLMs closer to the data source, enabling timely updates, reduced communication latency, and enhanced privacy by minimizing raw data transmission. This is particularly beneficial in scenarios requiring low-latency decision-making, such as autonomous driving~\cite{yang2021edge}, industrial automation~\cite{pandey2024security}, and smart healthcare~\cite{lai2022healthcare,el2021secure}. Moreover, the advancement of high-speed, low-latency wireless technologies can further facilitate efficient model updates and enable large-scale collaborative training across geographically distributed nodes~\cite{javaid2024leveraging}.

However, integrating wireless networks into RL-based LLMs also presents several challenges, which can be broadly categorized into four key aspects: data, model, environment, and user. From the data perspective, transmitting privacy-sensitive information, such as user activity or browsing history, over wireless channels increases the risk of data breaches and manipulation~\cite{guo2024survey}. From the model perspective, the LLMs must adapt to fluctuating network conditions, handle limited computational resources at the edge, and mitigate malicious attacks that can manipulate learning processes~\cite{qiaoICC_2024knowledge}. Additionally, maintaining model consistency and convergence across decentralized and distributed training settings further complicates deployment~\cite{qiao2023cdfed,shi2023improving}. From an environmental perspective, wireless networks often face constraints such as limited bandwidth, computing resources, and energy efficiency~\cite{cao2017towards}, which can hinder real-time inference, delay model updates, and introduce trade-offs between performance and resource consumption. Finally, from a user perspective, optimizing LLMs for diverse user demands and usage scenarios requires adaptive mechanisms that balance personalization, fairness, and efficiency while mitigating potential biases and security threats~\cite{zhao2025fairness,zhu2024recommender}.


Building on the previous discussion, this paper presents a comprehensive survey on the synergy between wireless networks and RL-based LLMs, exploring their motivations, challenges, potential solutions, future directions, and societal impact. To begin, we examine the recent advancements in wireless networks, highlighting the challenges and potential solutions in meeting the growing demands of modern applications. Subsequently, we explore the advancements in LLMs, offering an in-depth analysis of key training technologies, with a particular focus on the RL strategies used in LLM training, along with the associated challenges and potential solutions. Moreover, we explore the motivations behind the mutual empowerment of wireless networks and RL-based LLMs, illustrating how their synergy can enhance each other's strengths and drive transformative advancements. We then examine the challenges of integrating advanced LLMs into wireless communication systems and incorporating the benefits of wireless networks into LLM models, followed by insights into effective solutions for both strategies. Furthermore, we discuss potential future directions for deploying LLMs in wireless network environments and introducing wireless communication systems into LLMs, highlighting how emerging technologies can help overcome current limitations. Finally, we explore the broad applications and societal impact of the synergy between these two domains, paving the way for the development of next-generation intelligent communication systems and the evolution of more powerful and adaptable LLMs. Overall, by examining the challenges of LLMs and wireless networks and their interconnections, we aim to inspire future research that will drive the advancement of both fields, ultimately leading to a transformative era that pushes the boundaries of state-of-the-art LLMs and wireless network technologies.

\subsection{Major Contributions}
This is the first comprehensive survey that investigates the synergy between wireless networks and RL-based LLMs in terms of motivation, techniques, and applications. Previous surveys have focused on LLM-powered wireless networks from various perspectives, such as resource management~\cite{boateng2024survey,adhikary2024holographic}, network architectures~\cite{friha2024llm,xu2024large}, and prompt engineering~\cite{zhou2024large,shao2024wirelessllm}. However, these prior works have primarily concentrated on the benefits of wireless networks derived from the powerful generalization capabilities of LLMs.  In contrast, in this survey, we explore both the mutual empowerment and the symbiotic relationship between wireless networks and LLMs, considering how they can enhance each other's capabilities and overcome the challenges of each domain. We begin by summarizing the challenges and advancements in both wireless communication and LLMs. Next, we explore the intersection of these two domains, focusing on their motivations and the associated challenges. Subsequently, we offer novel insights and potential solutions to address these challenges. In addition, we examine the applications arising from the intersection of these two domains. Finally, we look ahead to future directions and examine the societal impact of the interplay between these two domains. In summary, the main contributions of this paper are as follows: 
\begin{itemize} 
\item[$\bullet$] Recognizing that prior work has mainly focused on the role of LLMs in enhancing wireless networks, this survey shifts the focus towards how wireless networks can, in turn, support LLMs. Meanwhile, we revisit the role of LLMs in optimizing wireless networks to provide a more comprehensive understanding.
\item[$\bullet$] Building on a comprehensive review of the key enabling technologies in both domains, we further analyze their synergy by identifying core motivations, major challenges, and potential solution strategies from two complementary perspectives: wireless networks for LLMs and LLMs for wireless networks.
\item[$\bullet$] Grounded in this dual-perspective analysis, we discuss future directions, emerging applications, and potential societal impacts resulting from the mutual empowerment between wireless networks and LLMs.
\end{itemize}

\subsection{Organization of This Paper}
The remainder of this survey is organized as follows. Section~\ref{sec:bridge_wireless_llm} provides a brief overview of the intersection between wireless networks and LLMs. Specifically, Sections~\ref{subsec:wireless_networks} and~\ref{subsec:rl_llms} introduce the key technologies behind each domain, while Section~\ref{subsec:wireless_meet_llm} briefly introduces their mutual benefits. Section~\ref{sec:empower_wireless_with_llms} explores how LLMs can enhance wireless networks, with Section~\ref{sec:motivation_empower_wireless_with_llms} outlining the motivations, Section~\ref{sec:challenges_empower_wireless_with_llms} discussing challenges, and Section~\ref{sec:solution_empower_wireless_with_llms} presenting potential solutions. Similarly, Section~\ref{sec:empower_llms_with_wireless} investigates how wireless networks can empower LLMs, with Section~\ref{sec:motivation_empower_llms_with_wireless} covering motivations, Section~\ref{sec:challenges_empower_llms_with_wireless} addressing challenges, and Section~\ref{sec:solutions_empower_llms_with_wireless} proposing potential solutions. In addition, Section~\ref{sec:applications} explores applications arising from the mutual empowerment of wireless networks and RL-based LLMs. Section~\ref{sec:directions} explores future research directions, while Section~\ref{sec:impact} examines societal impacts. Finally, Section~\ref{sec:conclusion} concludes the survey. Fig.~\ref{fig:paper_structure} illustrates the outline of this paper.

\begin{figure*}[t]
    \centering
    \includegraphics[width=\linewidth]{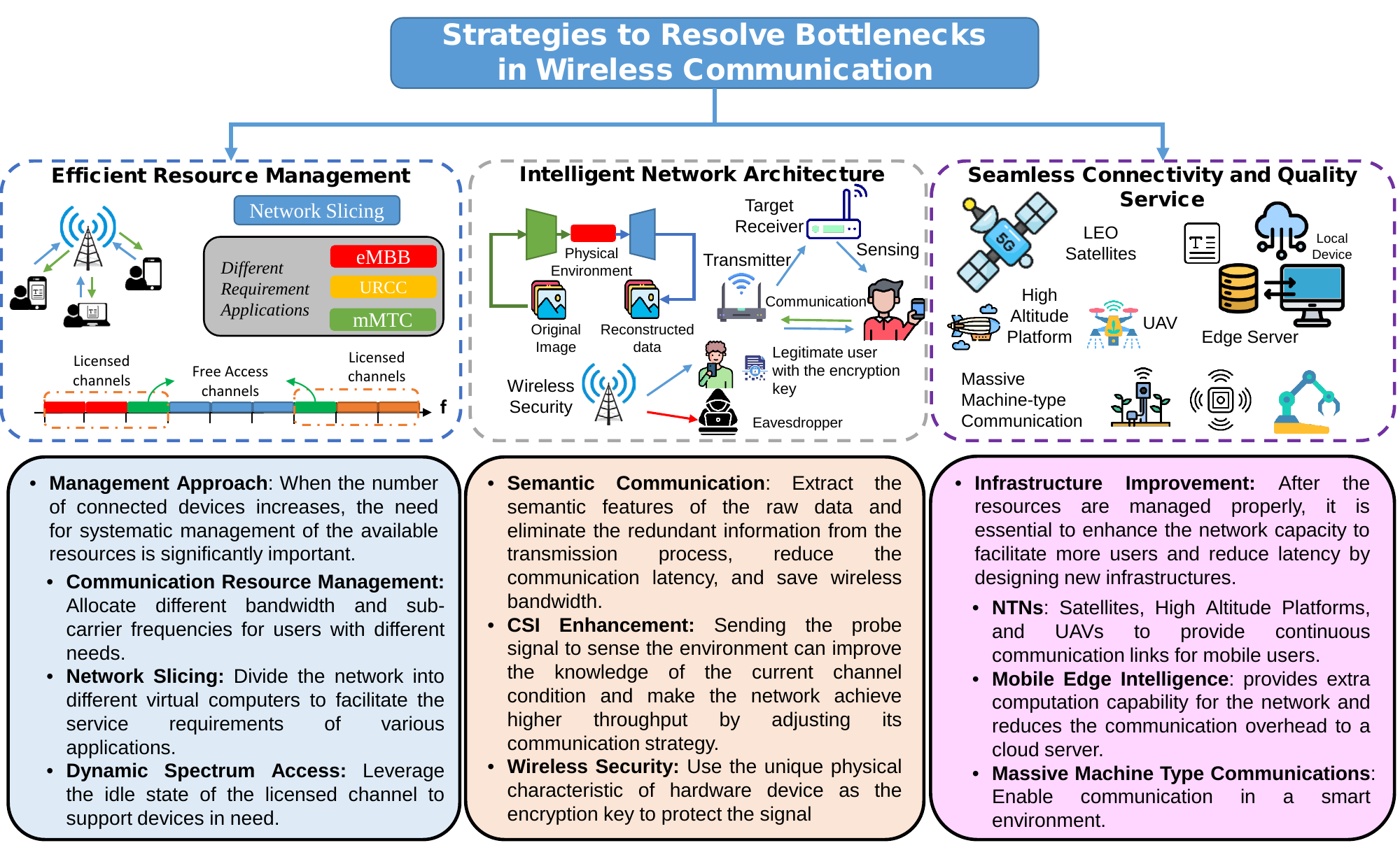}
    \caption{Various strategies for enhancing wireless communication systems: optimizing resource management in existing networks, improving transmission efficiency by transitioning from traditional to semantic communication, and leveraging environmental sensing to acquire physical context information for enhanced antenna beamforming.}
    \label{fig:wireless}
    \vspace{-7px}
\end{figure*}


\section{Bridging Wireless Networks and Large Language Models}
\label{sec:bridge_wireless_llm}

\subsection{Key Technologies for Optimizing Wireless Networks}
\label{subsec:wireless_networks}
The volume of data transmitted through wireless communication has surged exponentially in recent years, driven by several factors: the growing number of connected devices, the demand for high-quality data, and the increasing need for hyperconnectivity~\cite{loc2024semantic_communication,mozaffari2018communications}. With the emergence of advanced applications such as the Metaverse~\cite{sami2024metaverse}, 3D online gaming~\cite{wiboolyasarin2024digital}, telemedicine~\cite{lu2024promote}, and autonomous vehicles~\cite{parekh2022review}, wireless communication systems have been facing increasing challenges in meeting user demands due to the massive data requirements of these high-tech applications~\cite{aceto2019survey}. To address these challenges, advancements in wireless technologies can be broadly categorized into three complementary areas: (1) efficient resource management, (2) intelligent network architecture, and (3) integration of computing and communication. The resource management approach focuses on optimizing existing infrastructure to accommodate the growing number of devices, employing techniques such as dynamic spectrum allocation~\cite{8759030}. This method maximizes bandwidth efficiency by allocating more resources to active devices while reducing or suspending allocation for idle ones~\cite{9146125}. The intelligent network architecture approach explores new transmission paradigms, AI-driven optimization, and security enhancements to improve network performance~\cite{10417099}. Lastly, the integration of computing and communications approach investigates cutting-edge physical technologies, including higher-frequency communication, advanced antenna systems, and emerging innovations such as reconfigurable intelligent surfaces (RIS)~\cite{10137638}. A visual overview of these three solutions is presented in Fig.~\ref{fig:wireless}.

\subsubsection{Efficient Resource Management}
Within the network management approach, various techniques have been explored to accommodate the growing number of connected devices~\cite{cho2010mimo,8685766,8320765}. These include advanced modulation and multiple access techniques~\cite{dai2015non,6692652}, resource optimization strategies~\cite{10770127,10137638}, software-defined networking (SDN)~\cite{8320765}, and hybrid network architectures~\cite{fu2024iov}. Each individual technique has its advantages and can complement others in a unique way to achieve an efficient wireless communication framework. 

\textbf{Communication Resource Management.} Orthogonal Frequency Division Multiplexing (OFDM) 
\cite{OFDM,cho2010mimo,stuber2004broadband} and non-orthogonal multiple access (NOMA)~\cite{6692652,dai2015non,7230246} represent two prominent yet contrasting approaches. OFDM partitions the bandwidth spectrum into subcarriers, allowing wireless networks to transmit data without interference. In contrast, NOMA enables multiple users to share the same frequency by leveraging power-domain and code-domain techniques for signal separation. Another advancement in this approach is network slicing, which involves partitioning the network into several logical segments to support distinct application scenarios, such as enhanced mobile broadband (eMBB)~\cite{9146125}, ultra-reliable low-latency communications (URLLC)~\cite{9146125}, and massive machine-type communications (mMTC)~\cite{dutkiewicz2017massive}, rather than relying on a one-size-fits-all solution~\cite{8685766}. \textcolor{black}{Reserving bandwidth exclusively for certain applications such as television broadcasting, satellite communication\cite{Satellite}, and aeronautical can lead to inefficient use of wireless communication resources. To address this issue, dynamic spectrum access (DSA) has been introduced as a way to make use of underutilized spectrum without interfering with primary users \cite{4205091}. It works by actively identifying idle licensed channels and temporarily assigning them to secondary users. With the help of smart allocation algorithms, DSA offers a promising solution to the growing challenge of spectrum scarcity \cite{9759480}.}
    
\textbf{Network Slicing.} Network slicing enables the network to adapt to users' changing requirements~\cite{lei2022dynamic} dynamically. Note that the underlying concept of network slicing is virtualization, which decouples the network from its hardware dependencies and leverages available resources to run software functions~\cite{8320765}. In this context, virtual machines provide a platform for guest operating systems, enabling applications and services to be executed effectively~\cite{8759030,9146125}.

\subsubsection{Towards Intelligent Network Architecture Design: The Role of Semantic Communication and ISAC}
To meet the demands of increasingly complex applications and the explosive growth of connected devices, future wireless networks must evolve beyond rigid, manually managed infrastructures. Intelligent and programmable architectures offer enhanced adaptability, efficiency, and context-awareness, enabling the network to respond dynamically to complex and changing environments. Achieving this vision requires a fundamental redesign of existing network architectures, moving away from traditional network architecture towards deeply integrated and AI-empowered frameworks~\cite{yang2023witt,xu2024semantic,10531769}. Among them, semantic communication and integrated sensing and communication (ISAC) have emerged as representative innovations that significantly reshape the communication paradigm at both architectural and functional levels. 

\textbf{Semantic Communication.} Traditional communication systems perform well in low-noise environments, but communication failures occur when channel noise exceeds a certain threshold~\cite{trewavas2011information}. Semantic communication has emerged as a key technology for the next generation of wireless systems, as it focuses on the semantic meaning of the data being transmitted, eliminating redundant information from the transmitted signal to enhance efficiency~\cite{raha2023artificial,raha2023generative,luo2022semantic}. By jointly optimizing the source and channel encoders, semantic features are extracted from the raw data and compressed based on the channel conditions. This approach enhances robustness against physical noise and mitigates the "cliff effect"~\cite{getu2025semantic}. The rise of deep learning has enabled semantic communication to distinguish meaningful information from redundant data, significantly reducing the number of symbols transmitted and saving bandwidth resources for other users~\cite{8723589,9955525,hassan2025enabling,yang2023witt}. While the deep joint source-channel coding approach for semantic communication primarily focuses on data recovery, generative semantic communication offers greater flexibility by allowing the reconstructed data at the receiver to differ from the original~\cite{grassucci2024generative,xu2024semantic,nguyen2025contemporary,fan2024semantic}. However, this does not imply that the two sets of data are entirely different; they still share the same semantic meaning. Specifically, in tasks such as image retrieval, instead of transmitting full-sized image data, generative AI-based semantic communication can convert the images into descriptions. The receiver then uses these text descriptions to retrieve the corresponding images~\cite{yang2024rethinking}. Additionally, the advantage of generative AI models lies in their ability to process multimodal data, enhancing service quality while maintaining low communication costs~\cite{10531769}. 


\textbf{Integrated Sensing and Communication.} ISAC~\cite{10750351,10757511} offers an innovative solution to address the wireless communication bottleneck by utilizing the same spectrum for both sensing and communication, thereby eliminating the latency associated with channel switching, as well as the same hardware (antennas and transceivers). Through radar sensing, ISAC enables active beamforming to achieve optimal throughput and reduce communication latency. Additionally, with signal probing, ISAC provides high-accuracy environmental sensing data, which enhances channel state information (CSI) estimation. The improved CSI accuracy leads to better modulation and coding, ultimately increasing throughput and alleviating wireless traffic~\cite{10137638}. In addition, ISAC not only provides real-time environmental conditions but also tracks movement with high precision, making it highly beneficial for proactive beam steering and handover management~\cite{liu2022survey,9755276,10462480}. Another advantage of user trajectory prediction is the minimization of inter-user interference through optimized resource allocation strategies~\cite{10770127,10804189,10770161}. This technique offers various benefits on its own and can be integrated with other advanced technologies, such as mobile edge computing (MEC) and semantic communication, to drive further advancements~\cite{10757511,10007643,10417099,10750351,yang2024joint}. In addition to ISAC, numerous other approaches, such as channel estimation, also exist to enhance the quality of communication links~\cite{9919942,10575676,soltani2019deep}.

\textcolor{black}{\textbf{Wireless Security.} The explosive increase in the number of connected devices induces security threats to the users, as eavesdroppers and jammers have a higher chance of interfering with the communication link due to the density of the wireless network \cite{zhao2025wireless}. To address this, various types of wireless security have been studied, such as the physical layer, network layer and application layer, in which the physical layer security (PLS) has drawn the attention of the research community due to its massive potential. Specifically, PLS takes advantage of the unique physical characteristics of hardware devices and channels for authentication \cite{afeef2022physical}. Additionally, the development of AI technologies has been integrated into PLS, which successfully resolves the complexity problem of the environment and channels. In the work \cite{zhao2024generative}, the authors have demonstrated the benefits and drawbacks of traditional AI; then, they illustrate the merit of the adoption of generative AI for a secure physical layer.}

\subsubsection{Towards Seamless Connectivity: The Role of NTNs and MEC}
To meet the growing demands of high-quality service and seamless connectivity, modern wireless networks are undergoing a shift towards the next-generation computing and communication paradigm~\cite{hu2015mobile,liu2021reconfigurable}. This shift aims to enable more intelligent, efficient, and context-aware services across diverse environments. Key technological enablers of this trend include non-terrestrial networks (NTNs)~\cite{10004947}, which offer ubiquitous connectivity beyond terrestrial infrastructure, and mobile edge computing (MEC), which brings computation closer to the user to reduce latency and alleviate core network congestion. Additionally, reconfigurable intelligent surfaces (RIS) enhance signal propagation by dynamically shaping the wireless environment, while emerging high-frequency communication technologies such as millimeter-wave (mmWave) and terahertz (THz) enable ultra-high-speed data transmission. Massive multiple-input multiple-output (MIMO)~\cite{cho2010mimo}, leveraging large antenna arrays, further contributes to this integration by supporting concurrent high-throughput links for numerous users. Collectively, these technologies form the foundation of a new network paradigm in which computing and communication are no longer isolated components, but deeply intertwined to support future intelligent services.

\textbf{Non-Terrestrial Networks.} The existing wireless communication infrastructure relies heavily on fixed base stations, which struggle to deliver high-quality services in rural areas and special scenarios such as natural disaster response and military operations~\cite{imran2024exploring}. To meet the growing demand for ubiquitous connectivity, NTNs are designed to provide coverage across vast areas~\cite{nguyen2024semantic,10004947,9941481}. These networks can be deployed using unmanned aerial vehicles (UAVs)~\cite{10004947}, high-altitude platforms (HAPs)~\cite{tozer2001high}, or satellites operating in low-Earth orbit (LEO)~\cite{talgat2024stochastic}, medium-Earth orbit (MEO)~\cite{marshall2021investigation}, and geostationary orbit (GEO)~\cite{talgat2024stochastic}. The NTNs can be established by deploying a single type of these devices or a combination of multiple types, each offering distinct advantages. UAVs offer flexibility for the network based on ease of flight path adjustment, while satellites significantly expand coverage and offer greater resources in terms of computing power and energy compared to UAVs. With the moving property, the NTNs can create communication connections to remote users beyond the reach of fixed base stations. However, the proposal of NTNs, along with the MEC, increases the complexity of the wireless communication infrastructure, and it needs to be managed efficiently to be able to exploit the proposal's capabilities fully~\cite{10579820,7956189,8016573}. 

\textcolor{black}{\textbf{Massive Machine Type Communications.} The concept of smart factories and smart cities has flourished in recent years with the assistance of automated devices such as sensors, tracking devices, meters, and actuators. Communication between these devices is typically carried out among intelligent machines with minimal human intervention—or ideally, none at all—which is referred to as machine-type communications (MTC) \cite{dutkiewicz2017massive}. On the other hand, “massive MTC” refers to scenarios where the number of connected devices is at least ten times greater than current levels, with varying quality-of-service (QoS) requirements imposed on the cellular network \cite{dawy2016toward,7565189}. The technique has paved the way for the smart environment and is expected to be the pillar of next-generation communication \cite{8712527}. However, the key open questions about large-scale MTC deployment over cellular networks are when it will occur and how it should be designed and planned.}

\textcolor{black}{\textbf{Beamforming for mmWave and THz Communication.} With the large bandwidth offered by mmWave (30-300 GHz) \cite{8373698} and THz (0.1-10 THz) \cite{AKYILDIZ201416} communication, the wireless system can facilitate more users with ultra-high-speed data and low interference. These technologies can address the spectrum scarcity associated with the exponential growth of connected devices. However, since the wavelength of both techniques is significantly short, the short wavelength can be easily reflected, absorbed, or blocked by obstacles in the communication link. Therefore, it can only operate within short-range distances, and this is where the beamforming technique steps in to provide steering ability and create narrow beams with concentrated power toward the user, thus improving SNR and enabling a more extended communication range \cite{10045774,10845820}. Additionally, in a multiuser scenario, beamforming can achieve secure communication by preventing both active and passive eavesdroppers in the wireless transmission process with an appropriate resource allocation algorithm \cite{6781609}.}

\begin{figure*}[t]
    \centering
    \includegraphics[width=0.97\linewidth]{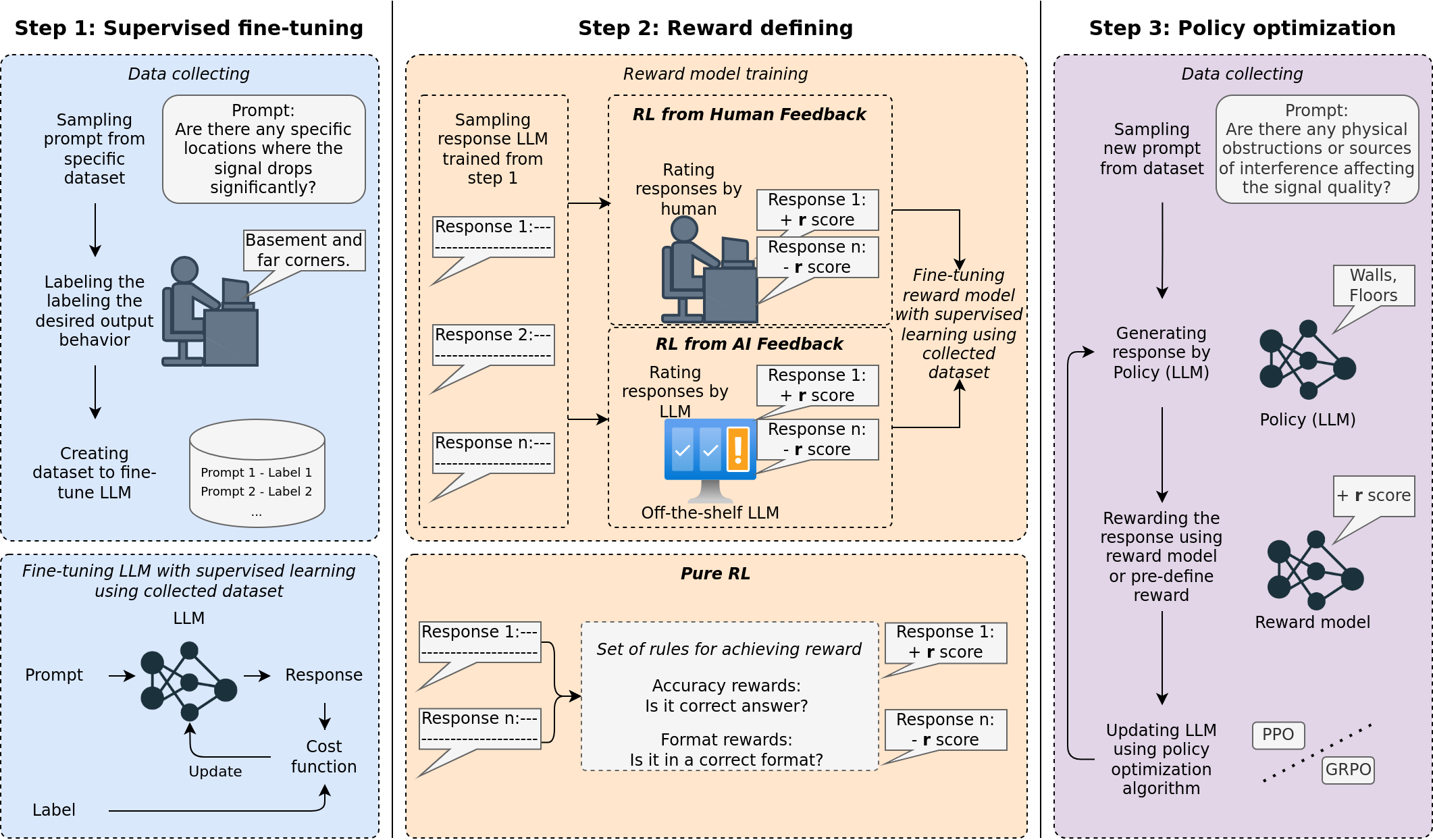}
    \caption{Diagram depicting the three stages of integrating RL into training LLMs: (1) Supervised fine-tuning, (2) Reward definition, and (3) Policy optimization~\cite{schulman2017proximal_arxiv, shao2024deepseekmath_arxiv}. Step 2 shows the comparison of three reward model definition approaches: RL from Human Feedback (RLHF)~\cite{ouyang_long2022RLHF}, RL from AI Feedback (RLAIF)~\cite{bai_yuntao2022RLAIF}, and Pure RL~\cite{guo_daya2025deepseekr1}. RLHF and RLAIF share similar pipelines, differing only in how they generate rating responses.}
    \label{fig:rl-based_overview}
    \vspace{-5px}
\end{figure*}

\textbf{Mobile Edge Intelligence.} MEI~\cite{hu2015mobile} is an emerging technology that enhances network performance by providing computational capabilities closer to end-users. These devices are widely distributed and positioned closer to users than cloud servers, thereby reducing the communication bottleneck for cloud servers~\cite{qiao2023mp}. By deploying computing resources closer to end-users, MEI minimizes communication delays and alleviates network congestion, making it particularly suitable for real-time applications~\cite{qiao2024logit}. Additionally, by decoupling the data and control planes into individuals, the SDN~\cite{haque2016wireless} can control the network effectively by adjusting the traffic routing, spectrum allocation to multiple access points, and slicing networks for different applications. Similar to MEI, another emerging technology in wireless communication is RIS~\cite{liu2021reconfigurable,huang2019reconfigurable}. RIS is designed to establish an indirect communication link between the transmitter and receiver by dynamically controlling signal reflection. This technology is particularly beneficial in environments where direct line-of-sight communication is obstructed~\cite{10306287,10458888}. RIS has been deployed in various scenarios, including smart buildings, vehicular networks, UAVs, and high-rise structures~\cite{park2024design,10154352,hassan2024optimizing}. By intelligently adjusting reflection coefficients, RIS can optimize signal propagation, reduce interference, and improve wireless link reliability. The integration of RIS with MEI and SDN further enhances network adaptability, providing a promising solution for future wireless communication systems. In addition to the substantial efforts dedicated to optimizing wireless networks, the rapid advancement of large language models (LLMs) has demonstrated significant potential in enabling AI-driven solutions for intelligent network management and optimization~\cite{xu2024large,xu2025serving}. The following subsection first introduces the core technologies underlying LLMs, with representative examples such as DeepSeek, and then explores the emerging synergy between wireless networks and LLMs.

\subsection{Key Techniques Behind DeepSeek and Beyond}
\label{subsec:rl_llms}

The emergence of LLMs has unlocked immense potential for AI-driven innovations in wireless communication~\cite{shao_jiawei2024decoder_application}. These models play a transformative role in network design, security, optimization, and management by leveraging their advanced capabilities in data analysis, pattern recognition, and decision-making. LLMs are applied across various aspects of wireless communication~\cite{zhang_ruichen2024wireless_application,qiu_kehai2024wireless_application,jiang_feibo2024wireless_application_arxiv}, including traffic allocation, failure detection, interference mitigation, and resource optimization, leading to substantial enhancements in network efficiency and reliability. LLMs can be categorized into three common architectures: encoder-only, encoder-decoder, and decoder-only, stemming from the vanilla transformer model. While encoder-only models such as BERT~\cite{devlin_jacob2019bert} and DeBERTa~\cite{he2021deberta} are proficient in analyzing network data logs and swiftly identifying irregularities, encoder-decoder architectures such as UL2~\cite{tay2022ul2_ICLR_arxiv} and ChatGLM~\cite{glm2024chatglm} are particularly effective for tasks requiring sequence-to-sequence transformations. These tasks include protocol conversion, optimization of signal processing, and enhancement of real-time communication capabilities~\cite{verma2024towards_arxiv,loc2024semantic_communication}. On the other hand, decoder-only architectures, such as DeepSeek-V3~\cite{liu2024deepseek}, GPT-4o~\cite{hurst_aaron2024gpt4o}, are good at predictive modeling and automated troubleshooting. These models are highly effective at power allocation, spectrum sensing, and protocol understanding~\cite{shao_jiawei2024decoder_application}. Their recent success is attributed to the efficient utilization of reinforcement learning (RL) techniques in training the LLMs. RL techniques, including policy gradient methods~\cite{wang2022policy_arxiv}, proximal policy optimization (PPO)~\cite{schulman2017proximal_arxiv} and group relative policy optimization (GRPO)~\cite{shao2024deepseekmath_arxiv}, play a crucial role in optimizing LLMs' decision-making processes in dynamic network environments. RL-based training strategies for LLMs can be broadly categorized into three main approaches: RL with human feedback (RLHF)~\cite{ouyang_long2022RLHF}, RL with artificial intelligence feedback (RLAIF)~\cite{bai_yuntao2022RLAIF}, and Pure RL~\cite{guo_daya2025deepseekr1}. Each of these methods offers distinct advantages in the training of LLMs for both general LLM tasks and more specialized wireless communication tasks~\cite{xu2024large,xu2025serving}. An overview of these approaches is illustrated in Fig.~\ref{fig:rl-based_overview}.

\begin{wrapfigure}{r}{0.50\textwidth} 
    \centering
    \includegraphics[width=\linewidth]{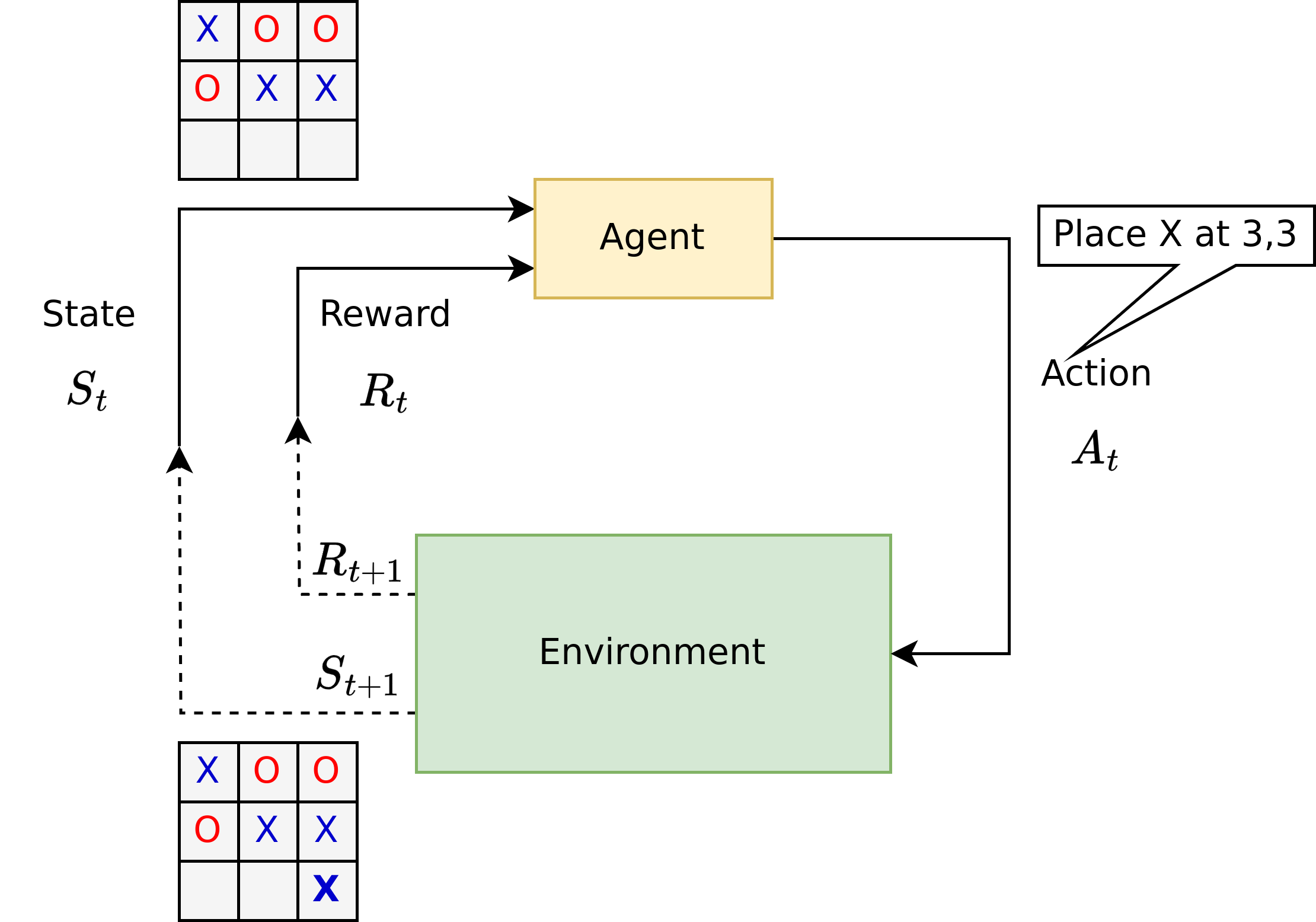}
    \caption{An example of an RL application in the Tic-Tac-Toe game. The agent learns optimal strategies through self-play, updating the state-action value function to enhance decision-making over time.}
    \label{fig:RL_tictactoe}
\end{wrapfigure}

\subsubsection{Reinforcement Learning}

RL stands out as a promising research area within machine learning that has gained substantial attention in recent years~\cite{wang2024reinforcement,milani2024explainable,lei2020deep}. RL tasks involve learning how to solve a problem by taking a series of actions aimed at maximizing cumulative rewards~\cite{sutton2018reinforcement}. 
To illustrate, we consider the classic game of Tic-Tac-Toe as shown in Fig.~\ref{fig:RL_tictactoe}, where the objective is to form a row, column, or diagonal within a 3x3 grid before your opponent does. Success in this game hinges on making strategic moves (placing X's or O's in the grid), with each move influencing the outcome of subsequent actions. Imagine the player in this scenario as an AI agent; our goal is to train this AI agent to win the game. While traditional AI approaches, such as supervised and unsupervised learning, may struggle with this task due to its complexity, RL offers a fresh perspective. In RL, the problem is viewed through the lens of five key components: agent, environment, agent action, agent state, and rewards. Fig.~\ref{fig:RL_tictactoe} illustrates how these components interact to win the Tic-Tac-Toe game. In this approach, the agent acts based on the current situation, interacts with the environment to get feedback, and repeats this process until a winner is determined in each game, known as an episode in RL. The agent learns to improve its strategy by trial and error, aiming to maximize the rewards received in each episode. In natural language processing (NLP), we can view the agent as our LLM. The agent generates an action (response) based on its current state (prompt) and policy (LLM) to interact with the environment. The environment, in turn, responds to the agent's action through rewards and transitions the current agent state to the next agent state (next prompt). Depending on our training objectives, the environment needs to be defined to provide rewards based on agent actions and states. This approach aims to maximize the number of rewards generated by the environment through our agent action. Specifically, in the context of LLM, defining the reward signal is crucial for optimizing the model using the RL approach. The difference in constructing reward models will be presented in the next section.

\subsubsection{RLHF-based LLMs}
RLHF~\cite{ouyang_long2022RLHF} is an optimization method that integrates RL techniques with the reward model trained from the human feedback dataset to enhance the LLM's ability to interpret and act according to the user's intentions. By incorporating this human-generated reward model into the RL loop, RLHF strives to minimize the occurrence of inaccurate or uninformative outputs from the LLM. The common process of RLHF involves three common steps: supervised fine-tuning, reward model training, and policy optimization.

\textbf{Supervised Fine-Tuning.} This step focuses on refining LLMs such as DeepSeek-V3~\cite{liu2024deepseek} and GPT-4o~\cite{hurst_aaron2024gpt4o}, transitioning them from a broad domain to one tailored for specific tasks. It involves collecting demonstration data and training a supervised policy to establish a robust foundation LLM for subsequent RL processes, as shown in step 1 of Fig.~\ref{fig:rl-based_overview}. Similar to the cold start phase of DeepSeek-R1~\cite{guo_daya2025deepseekr1}, this step aims to mitigate initial instability in RL training that may arise from the base model. The demonstration dataset includes pairs of prompt inputs and desired output labels provided by humans, which showcase the desired output behavior. For example, a prompt input might be "Are there any specific locations where the signal drops significantly?" and the corresponding desired output would be the accurate locations where the signal is dropped. This data guides the model in learning the correct responses and behaviors for specific tasks.

\begin{wrapfigure}{r}{0.5\textwidth}
    \centering
    \includegraphics[width=0.95\linewidth]{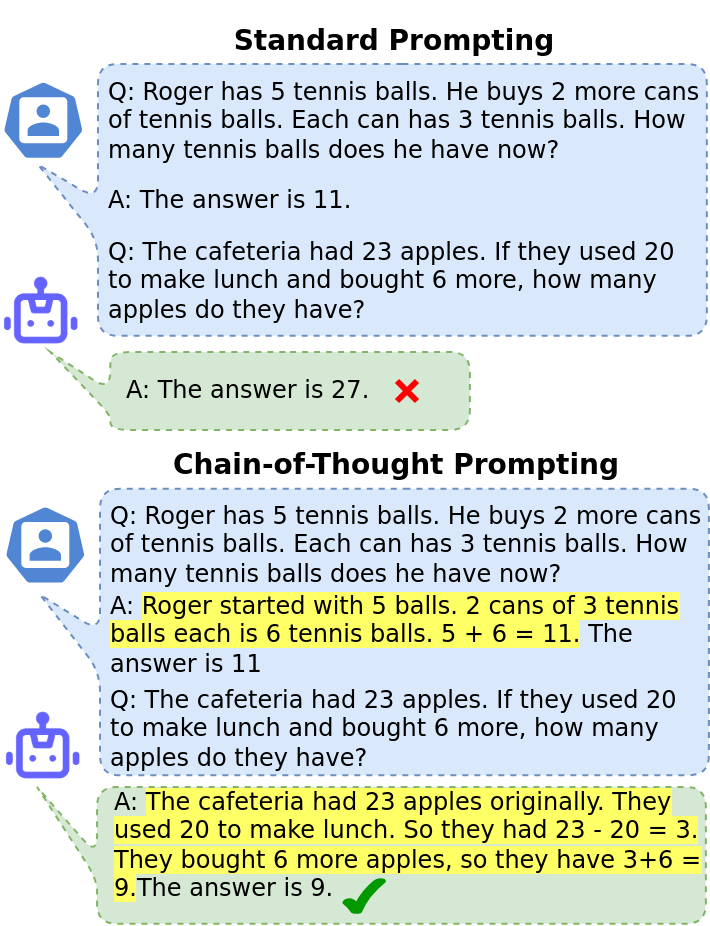}
    \caption{An example of CoT. Different from standard prompting, CoT prompting explicitly outlines the reasoning process, leading to more interpretable and accurate results.}
    \label{fig:CoT}
\end{wrapfigure}

\textbf{Reward Model Training.} The reward model serves as the cornerstone of RL within LLM frameworks, providing a training signal for updating the model and guiding it toward desired behaviors. In the context of RLHF, this reward model assesses the LLM's outputs by assigning them scalar values based on their quality. The primary objective of training in this framework is to optimize the reward obtained from the reward model for each interaction. To train the reward model, RLHF initially gathers outputs from multiple LLMs based on a given prompt. These outputs are then evaluated by a human labeler who ranks them according to specific criteria, ranging from best to worst. Subsequently, a supervised learning approach is employed to train the reward model, leveraging the labeled data to refine the model's ability to evaluate and score the LLM outputs accurately, as shown in step 2 of Fig.~\ref{fig:rl-based_overview}. Several notable approaches, such as InstructGPT~\cite{ouyang_long2022RLHF} and GPT-4~\cite{achiam2023gpt}, adopt this strategy to guide their models effectively. By incorporating human feedback into the training process and iteratively refining the reward model's evaluation capabilities, these advanced language models strive to produce outputs that better align with user expectations and preferences.

\textbf{Policy Optimization.} In RL, policy optimization shares similarities with model optimization in deep learning, as both involve updating model weights. However, RL differs in that the model does not receive immediate feedback at each step to indicate whether the update direction is correct. Instead, the current policy produces outputs that are later assessed by a reward model, generating corresponding reward scores. This delayed feedback mechanism, known as "delayed rewards" in RL, means the model only receives evaluations after completing text generation or a sequence of actions (steps). Optimization in RL is commonly performed using methods like PPO~\cite{schulman2017proximal_arxiv} or GRPO~\cite{shao2024deepseekmath_arxiv}. PPO, a widely adopted approach in RL, ensures stability during policy updates but requires both policy and value models, which can lead to high computational costs for large-scale tasks. In contrast, GRPO introduces a novel group-relative optimization framework that emphasizes relative performance within learning groups rather than absolute metrics, reducing computational overhead while maintaining effectiveness.

RLAIF~\cite{bai_yuntao2022RLAIF} follows a similar pipeline to RLHF, with the key distinction being that RLAIF offers an efficient approach to training a reward model by leveraging existing LLM. As illustrated in the reward model training block of step 2 of Fig.~\ref{fig:rl-based_overview}, RLAIF differs from RLHF by replacing human feedback labels with a process that involves presenting the same task to an independent, pre-trained LLM. RLAIF achieves this by selecting more harmless responses, such as multiple-choice options, generated within a predefined format that effectively conveys the target principles for the given task. Subsequently, log probability responses are computed to derive scalar scores, which serve as essential signals for the reward model's training and optimization. In addition, varying the formats and principles employed in this process can lead to notably more robust model behavior. By adopting this strategy, the need for extensive human effort in dataset collection is reduced, as constructing datasets manually can be a time-intensive process. By utilizing RLHF or RLAIF, LLMs can be adjusted to exhibit the intended output behavior. However, these approaches pose challenges related to computational overhead and inference time, thereby impeding the training progress of RL~\cite{cao_yuji2024rl_llm_challenge}. Additionally, the process of gathering data labels for training the reward model is laborious and expensive for both RLHF and RLAIF, rendering these approaches impractical for adaptation to different domains.

\subsubsection{Pure RL-based LLMs}
\label{subsub:pure_rl}

Another innovative approach involves replacing the reward model with predefined reward rules, as proposed by DeepSeek-R1~\cite{guo_daya2025deepseekr1}, which has demonstrated exceptional performance in reasoning tasks for LLMs. This method eliminates the need for supervised fine-tuning, relying solely on RL to guide model optimization. Notably, recent findings from DeepSeek-R1-Zero provide the first evidence that reasoning behaviors can naturally emerge through pure RL~\cite{guo_daya2025deepseekr1}, highlighting its potential as a powerful alternative to conventional training paradigms. Specifically, instead of relying on a pre-trained reward model derived from a rating dataset, DeepSeek-R1 implements a rule-based reward system comprising two primary types of rewards: accuracy rewards and format rewards. Accuracy rewards are designed to enable the model to assess the correctness of its responses. For instance, in scenarios such as math problems with deterministic solutions, the model can only earn a point if it provides the correct answer. On the other hand, format rewards force the model to respond in accordance with a predefined format to be considered a valid response and earn a point. In addition, DeepSeek-R1 replaces the input prompt with a specific reasoning question format for training the model~\cite{guo_daya2025deepseekr1}. This technique allows LLMs to generate a chain of thought (CoT) that can explain the reasoning behind the answer. Fig.~\ref{fig:CoT} illustrates an example of the reasoning process. Prior to providing the answer, LLMs engage in extensive logical deliberation, sometimes even self-correcting with an "aha moment"~\cite{guo_daya2025deepseekr1} before giving the final response. Notably, this technique helps train the reason model without necessitating collected data to train the reward model, thereby enhancing its scalability and adaptability across diverse contexts. However, this approach also has its drawbacks, notably in its low generalizability and performance compared to other methods in other tasks~\cite{guo_daya2025deepseekr1}. Moreover, the complexity of defining reward rules for specific domains presents a significant obstacle, hindering the seamless application of this approach across various domains. As a result, while pure RL approaches offer a promising and scalable alternative to traditional training pipelines, further research is required to improve their adaptability, robustness, and applicability to a wider range of real-world scenarios~\cite{xu2025dark,guo_daya2025deepseekr1}.

\subsection{Wireless Networks Meet LLMs}
\label{subsec:wireless_meet_llm}

Although significant progress has been made in optimizing wireless networks through advanced resource management and intelligent architectures, these solutions often rely on task-specific algorithms that lack generalization and adaptability~\cite{chen2024big}. In contrast, the emergence of LLMs, such as DeepSeek~\cite{guo_daya2025deepseekr1}, ChatGPT~\cite{achiam2023gpt}, and Gemini~\cite{team2023gemini}, has introduced a powerful new paradigm for cross-domain intelligence and decision-making. With their exceptional capabilities in understanding, reasoning, and multimodal interaction, LLMs have the potential to fundamentally transform how wireless networks are designed, managed, and optimized. In particular, LLMs enable learning from heterogeneous data sources, extracting insights from complex network states, and controlling network elements through multimodal inputs. Conversely, wireless communication infrastructures play a critical role in supporting the efficient development and scalable deployment of LLMs. This bidirectional relationship opens up promising opportunities for building LLM-driven wireless frameworks, while simultaneously leveraging wireless advancements to enable more intelligent and efficient LLM systems. Despite these emerging possibilities, there remains a lack of contemporary surveys exploring the mutual empowerment of these two rapidly evolving fields. Therefore, in the following section, we examine the synergy between wireless networks and LLMs from three perspectives: motivations, challenges, and potential solutions. This discussion lays the foundation for a deeper understanding of how the integration of LLMs can unlock new levels of intelligence and automation in future wireless systems, and conversely, how advances in wireless networks can enhance the capabilities and deployment of LLMs. It is worth noting that since most popular LLMs (e.g., ChatGPT, Gemini, Grok, and DeepSeek) incorporate RL in their training process, we use the terms RL-based LLMs and LLMs interchangeably throughout this paper.

\section{RL-based LLMs for Wireless Networks}
\label{sec:empower_wireless_with_llms}

\begin{wrapfigure}{r}{0.50\textwidth} 
    \centering
    \includegraphics[width=\linewidth]{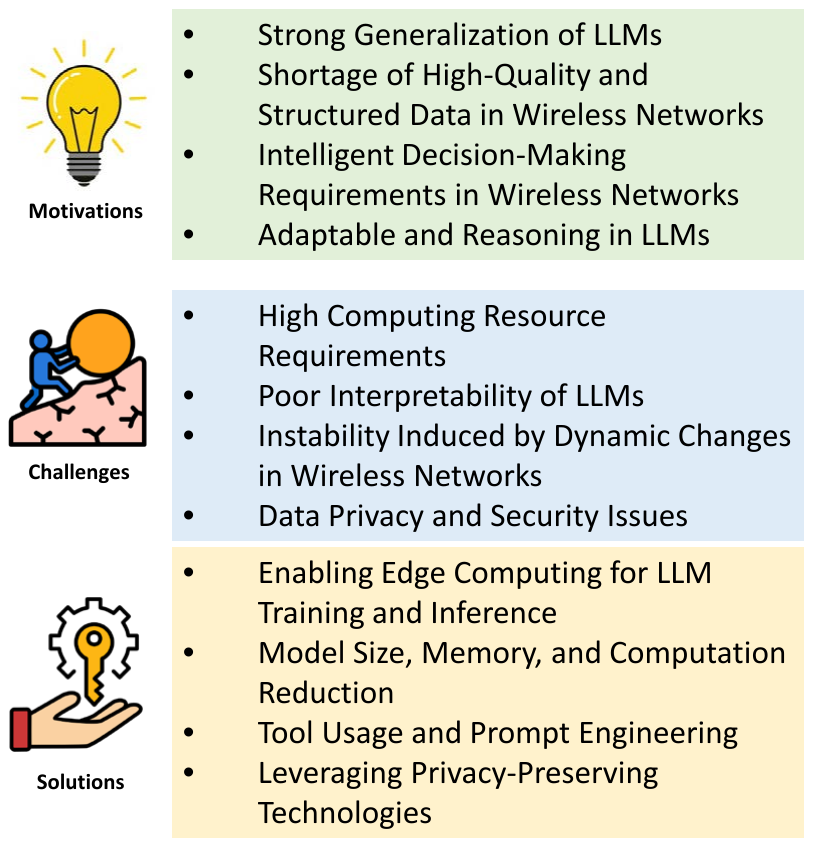}
    \caption{Motivations, challenges, and potential solutions for empowering wireless communication systems with LLMs.}
    \label{fig:llm_for_wireless}
\end{wrapfigure}

In this section, we outline the motivations for integrating RL-based LLMs into wireless communication systems, such as the strong generalization capabilities of LLMs, the challenges posed by the scarcity of high-quality structured data in wireless networks, and the potential for improved decision-making in dynamic environments. We then discuss the challenges faced by wireless communication systems when integrating RL-based LLMs, such as the large computational resource requirements and the poor interpretability of LLMs.
Finally, we discuss potential solutions to these challenges, such as leveraging edge computing for LLM training and inference, as well as optimizing model size, memory, and computational efficiency. Fig.~\ref{fig:llm_for_wireless} offers a comprehensive overview of the motivations, challenges, and potential solutions for integrating LLMs into wireless communication systems.

\subsection{Motivations for Empowering Wireless Networks with LLMs}
\label{sec:motivation_empower_wireless_with_llms}

\subsubsection{Strong Generalization of LLMs} Wireless networks operate in highly dynamic and complex environments where network conditions, user demands, and interference patterns fluctuate over time~\cite{letaief2006dynamic,mao2018deep}. For instance, network congestion can vary depending on the number of users in a specific area, while interference may be influenced by geographic location and environmental factors~\cite{adhikary2023transformer,adhikary2024power,wei2020assessing}. Similarly, user demands, such as data rates or latency requirements, can change throughout the day~\cite{jiang2018low}. In these situations, traditional machine learning models often struggle to adapt, requiring frequent retraining with newly collected data, which can be costly and inefficient. 

In contrast, LLMs possess strong generalization capabilities, enabling them to effectively manage diverse and evolving situations without the need for constant retraining~\cite{zheng2024towards}. Generalization refers to the ability to make effective decisions and predictions in unseen domains without requiring retraining~\cite{raha2024boosting}. This capability is enabled by leveraging knowledge from large-scale pretraining, which allows LLMs to transfer insights across different domains, thereby enhancing both efficiency and adaptability. For instance, an LLM trained on a wide range of textual data can be easily adapted to tasks such as medical diagnosis~\cite{shi2024medadapter}, sentiment analysis for social media~\cite{kumar2023analyzing}, or machine translation~\cite{eschbach2024exploring} with minimal fine-tuning. Therefore, this generalization ability is especially valuable in wireless networks, where conditions are constantly evolving, and the ability to adapt to new scenarios quickly is crucial for maintaining efficient and reliable performance.

\subsubsection{Shortage of High-Quality and Structured Data in Wireless Networks} Wireless networks often face the challenge of obtaining high-quality, structured data due to the dynamic nature of the environment and the decentralized distribution of data sources, particularly in the era of big data~\cite{sharma2017live,nguyen2020enabling}. For example, wireless communication systems rely on a wide variety of sensors, such as base stations, user equipment, and environmental factors like weather conditions, which produce vast amounts of unstructured and noisy data~\cite{dai2019big,9919976,garg2010wireless}. This data is often incomplete, inconsistent, and challenging to organize into a usable form for analysis or model training. Moreover, the vast amount of data generated across different devices and locations makes it difficult to collect and label the necessary datasets for traditional machine learning approaches.

LLMs, on the other hand, can potentially address this issue by leveraging their ability to process and make sense of unstructured data~\cite{vijayan2023prompt}. By pretraining on large-scale data, LLMs can extract meaningful patterns and relationships from vast amounts of raw, unstructured data without the need for explicit labeling or structured data~\cite{dagdelen2024structured}. For instance, LLMs can be trained on logs from wireless networks or communication traces, where they learn to identify patterns of network congestion, interference, and traffic fluctuations, even when the data is noisy or incomplete~\cite{kan2024mobile}. With minimal fine-tuning, LLMs can apply these insights to improve network resource allocation, optimize routing protocols, or predict future demand without requiring the extensive amount of structured, labeled data traditionally needed for such tasks. The proof-of-concept results from WirelessLLM~\cite{shao2024wirelessllm} highlight the effectiveness of LLMs in understanding and addressing real-world wireless communication scenarios, such as network monitoring, resource management, and protocol understanding.

This ability to work with noisy, unstructured data and still deliver reliable predictions makes LLMs a promising solution for the data scarcity challenges inherent in wireless networks. Moreover, LLMs can also generate synthetic data to supplement the scarcity of high-quality, structured data in wireless networks~\cite{long2024llms,tao2024harnessing}. By leveraging their generative capabilities, LLMs can simulate realistic network scenarios, creating data that reflects various conditions, such as different levels of interference, congestion, or traffic patterns. This synthetic data can be used to train models, test algorithms, or augment existing datasets, thereby overcoming data limitations and improving the robustness of wireless network systems.

\subsubsection{Intelligent Decision-Making Requirements in Wireless Networks} 
Wireless networks are subject to dynamic and unpredictable conditions, such as interference, congestion, and mobility of devices~\cite{mobile2002wireless,praveen2024link}. These constantly changing factors make it difficult to predict traffic patterns and optimize resources effectively. Meanwhile, effective resource management, such as spectrum allocation, power control, and user scheduling, is essential to ensure optimal network performance. However, these tasks require continuous real-time adaptation and decision-making, as network conditions change rapidly and are often difficult to predict. Traditional methods, such as heuristic-based user scheduling~\cite{zhong2019hyper}, rule-based scheduling~\cite{selvi2019rule}, and static power control algorithms~\cite{pantazis2007survey}, often fail to adapt to the dynamic nature of the network~\cite{liu2024survey}.

To overcome the limitations of traditional methods, which often suffer from unreliable decision-making, LLMs present a compelling alternative by leveraging their inference capabilities to process complex, unstructured data and make intelligent, context-aware decisions~\cite{qiu2024large}. For instance, an LLM can analyze both historical trends and real-time network data to predict congestion and proactively adjust resource allocation, such as reallocating spectrum or modifying transmission power to prevent network overload.~\cite{zhou2024large} demonstrates that leveraging the inference capabilities of LLMs can achieve performance comparable to traditional reinforcement learning while bypassing extensive model training and hyperparameter tuning. Moreover, when integrated with RL, LLMs enable dynamic optimization of resource management by incorporating historical data, real-time inputs, and human feedback to adapt to evolving network conditions. By continuously learning from human input and channel feedback, the system can update itself in real time, ensuring more responsive and adaptive network management.


\subsubsection{Adaptable and Reasoning in LLMs} Since LLMs are trained on diverse data sources, they demonstrate remarkable adaptability, enabling them to be customized for specific domains through techniques such as fine-tuning~\cite{xin2024parameter}, transfer learning~\cite{adhikary2024transfer,tan2018survey}, prompt engineering~\cite{sahoo2024systematic}, and retrieval-augmented generation (RAG)~\cite{yu2024evaluation}. These techniques enable LLMs to quickly become domain-specific experts, understand complex tasks, and respond effectively to specialized queries. Specifically, fine-tuning allows LLMs to optimize predictions based on domain-specific nuances in downstream tasks, making them highly effective in those contexts. Transfer learning enables LLMs to transfer knowledge from one domain or task to another, leveraging patterns and features learned from the source domain to adapt more rapidly to new tasks or domains with less data. Similarly, prompt engineering allows users to design tailored instructions that guide the behavior of the LLM, ensuring it produces relevant and precise responses within the target domain. Finally, the integration of RAG further enhances adaptability by enabling access to external knowledge sources, ensuring up-to-date and context-aware decision-making. This adaptability empowers the model to continuously adjust and optimize its performance across domains, minimizing the need for extensive retraining.

In addition to adaptability, the latest LLMs now possess advanced reasoning capabilities, such as chain of thought (CoT)~\cite{wei2022chain} methods found in models like DeepSeek-R1~\cite{guo_daya2025deepseekr1} and GPT-4o~\cite{hurst_aaron2024gpt4o}. The CoT method provides transparent and logical thought processes, enabling the models to break down complex problems into simpler components and reason through intricate scenarios step by step. For instance, CoT can explain why the sea appears blue by detailing how water molecules preferentially absorb longer wavelengths (e.g., red) while scattering shorter wavelengths (e.g., blue), resulting in the characteristic blue color of the ocean. By elucidating their reasoning chain, LLMs not only demystify challenging problems but also offer valuable insights for decision-making. This capacity to deconstruct complex tasks into understandable steps is particularly critical in managing the uncertainty and variability inherent in wireless networks. By simplifying complex tasks and making intelligent predictions without the need for extensive retraining, LLMs can optimize resource allocation and enhance network management, thereby providing effective solutions to the dynamic challenges of wireless environments.



\begin{figure*}[t]
    \centering
\includegraphics[width=\linewidth]{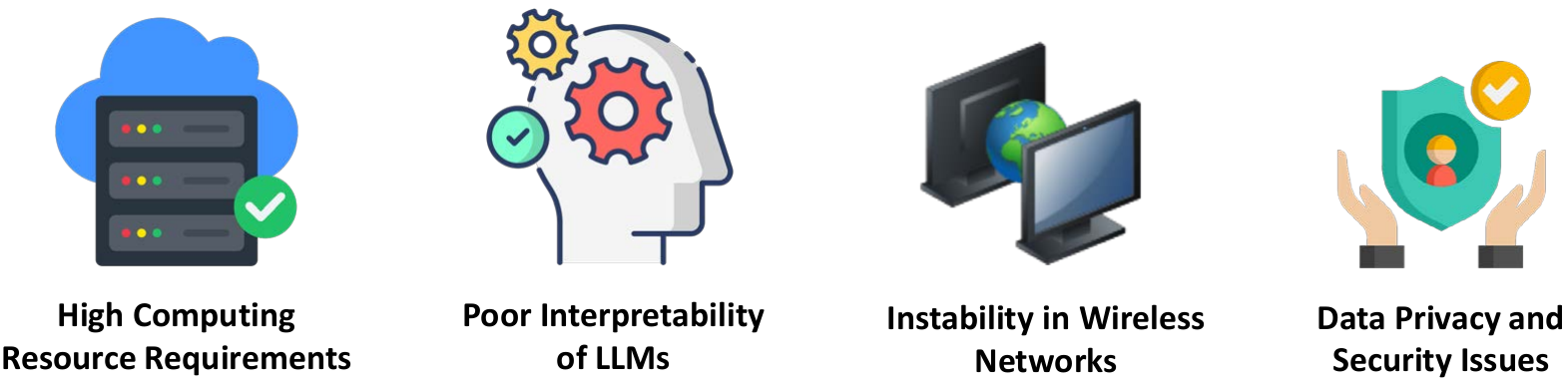}
    \caption{Challenges for empowering wireless communication systems with LLMs.}
\label{fig:llm_for_wireless_challenges}
\end{figure*}

\subsection{Challenges}
\label{sec:challenges_empower_wireless_with_llms}

While integrating LLMs into wireless networks can effectively tackle numerous issues in this domain, it also presents other challenges that need to be addressed. These challenges encompass aspects ranging from data considerations and model complexity to resource and hardware demands, as shown in Fig.~\ref{fig:llm_for_wireless_challenges}.

\subsubsection{High Computing Resource Requirements} LLMs are structured on transformer architecture, known for their quadratic complexity~\cite{keles2023computational}. Given that wireless networks necessitate ultra-low latency for communication, striking a balance between accuracy and complexity becomes imperative~\cite{sutton2019com_survey}. Researchers and developers are tasked with crafting efficient models that can navigate this challenge effectively. Moreover, the consideration of power consumption is a necessity, particularly for devices such as IoT or mobile devices. This additional constraint further complicates the optimization of models for real-world deployment within resource-constrained environments.

\subsubsection{Poor Interpretability of LLMs} Although there are efforts to develop tools and techniques for explaining and interpreting LLMs~\cite{yang2023survey}, these resources are still in their early stages of development. Existing methods, such as AttentionViz~\cite{catherine2024xai_att_viz} and feature visualization~\cite{yingfan2021xai_feature}, provide some insights but are often limited in fully explaining the model's behavior. Consequently, it is important to develop techniques that can explain how LLMs generate outputs, facilitating the regulation of these models' behavior for industrial applications. Moreover, these techniques can empower researchers to enhance their understanding of LLMs, enabling them to tackle limitations and boost performance effectively.

\subsubsection{Instability Induced by Dynamic Changes in Wireless Networks} The interactions among various parameters and layers in LLMs are non-linear, meaning that small changes in input can result in significant and unpredictable output changes~\cite{kumar2024large}. This makes it difficult to predict and interpret the model's behavior across diverse scenarios. The vast number of parameters and layers~\cite{guo_daya2025deepseekr1,hurst_aaron2024gpt4o, reyes2024evaluating} involved within these models poses a challenge in tracing and understanding their decision-making process. In wireless communication, this complexity is further compounded by the dynamic and heterogeneous nature of network environments and the need to process and analyze vast amounts of unstructured data available on most network devices~\cite{xu2021heterogeneous_network}. Additionally, the application of semantic communication~\cite{nguyen2024efficient}, which aims to understand and interpret the meaning behind the transmitted data rather than just the raw bits, introduces further challenges. When only the context is conveyed in the message, the inputs to LLMs may vary, yet they must produce consistent outputs based on this shared context.

\subsubsection{Data Privacy and Security Issues} LLMs require large datasets for training to build a robust model for wireless networks~\cite{guo_daya2025deepseekr1,hurst_aaron2024gpt4o, reyes2024evaluating}. However, these data samples may contain user-sensitive information that can not be leaked outside the network~\cite{yan2024protecting}. It is imperative to anonymize and secure this data during collection and distribution to forbid unauthorized access and potential abuse. Furthermore, LLMs can be vulnerable to exploitation, potentially exposing user information and offering adversaries fresh avenues for attack~\cite{das2025llm_security}. Therefore, guaranteeing the accuracy and integrity of the data employed by LLMs is significant in maintaining communication security.

\subsection{Potential Solutions}
\label{sec:solution_empower_wireless_with_llms}

\subsubsection{Enabling Edge Computing for LLM Training and Inference} Integrating edge computing~\cite{zhang2024llm_edge_computing} can enhance the training and inferencing processes in LLMs, resulting in a significant reduction in latency. This approach not only diversifies the dataset by incorporating different data sources but also speeds up response times from LLMs by bringing computation and data storage closer to data sources.  Moreover, local data processing on edge devices or devices close to edge devices can minimize the transmission of sensitive information over the internet, thereby enhancing data privacy and security~\cite{zhang2018data}. Additionally, leveraging edge computing can decrease operational expenses by reducing dependence on centralized cloud infrastructure and lowering data transfer costs~\cite{zhao2019edge}. This improves the scalability of LLM applications, enabling them to manage larger data volumes and more intricate tasks.

\subsubsection{Model Size, Memory, and Computation Reduction} LLMs are usually large and complex, but several techniques can make them smaller and faster without losing performance. Pruning~\cite{ashkboos2024slicegpt} involves cutting out less important parts of the model to reduce its size without much impact on how well it works. Quantization~\cite{wenqu_shao2024ICLR} converts high-precision data into lower-precision data, such as changing 32-bit numbers to 8-bit numbers, to save space while maintaining similar performance. Knowledge distillation~\cite{gu2024minillm} involves training a smaller model (student) to copy the behavior of a larger model (teacher), effectively transferring knowledge while reducing size. Caching~\cite{bang2023gptcache} stores intermediate results so they can be reused later, reducing the need for repeated calculations and speeding up the response times of LLMs. By implementing these strategies, LLMs can be integrated into wireless networks while mitigating the complexities associated.

\subsubsection{Tool Usage and Prompt Engineering} 
The challenge of interpretability in LLMs is generating accurate predictions supported by accurate explanations. One approach to address this issue is equipping LLMs with the capability to offer CoT prompt~\cite{wei2022chain}, enabling them to break down complex problems into simpler components and reason through intricate scenarios step by step. This CoT can be achieved through the utilization of RL-based techniques, as demonstrated in~\cite{guo_daya2025deepseekr1} and discussed in Section~\ref{subsub:pure_rl}. By integrating the CoT capability, LLMs can improve decision-making processes and enhance the reliability of their decisions. This transparency can enhance trust and understanding of the model's inner workings, increasing acceptance and utilization in various applications. Additionally, the ability of LLMs to break down complex problems and reason through them step by step can significantly benefit users in comprehending the rationale behind the model's outputs. Beyond CoT, more advanced prompting strategies, such as tree-of-thought (ToT)~\cite{yao2024tree} prompting and graph-of-thought (GoT)~\cite{besta2024graph} prompting, further enhance interpretability. ToT prompting structures reasoning as a tree, enabling LLMs to explore multiple paths, backtrack, and reassess solutions. GoT prompting extends this by using a graph-based approach, allowing flexible, non-linear problem-solving similar to human thinking.

Furthermore, leveraging external tools like the retrieval-augmented generation (RAG) mechanism and external validation sources allows LLMs to verify outputs and enhance information retrieval accuracy. This is expected to reduce hallucination risks while improving response trustworthiness~\cite{shao2024wirelessllm}. For example, integrating a mathematical solver can enable precise calculations, while code execution environments allow models to generate and verify programmatic outputs before returning a response. Carefully designed prompts can further reinforce step-by-step reasoning, ensuring that the model’s outputs are not only correct but also explainable.

\begin{figure*}[t]
    \centering
    \includegraphics[width=0.95\linewidth]{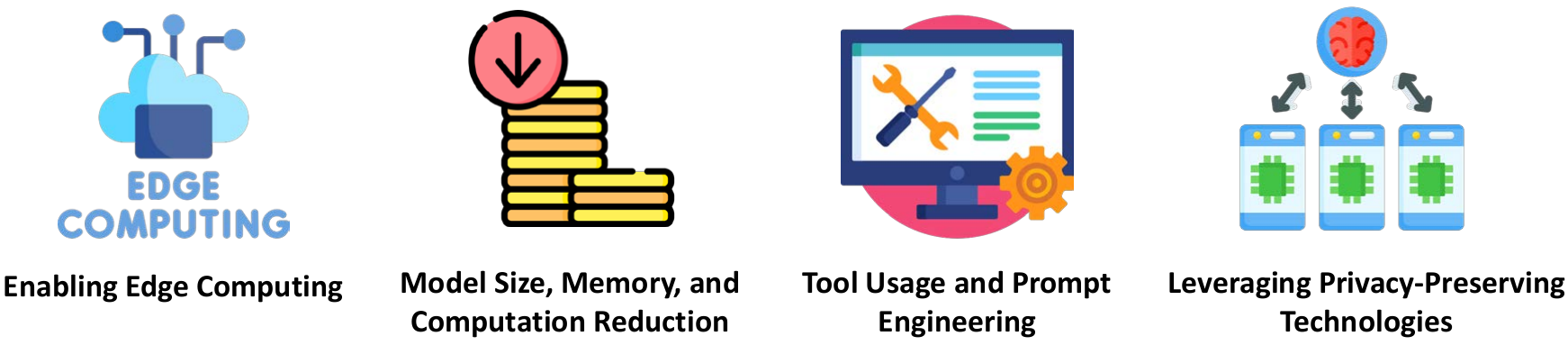}
    \caption{Potential solutions for empowering wireless communication systems with LLMs.}
    \label{fig:llm_for_wireless_solutions}
\end{figure*}

\subsubsection{Leveraging Privacy-Preserving Technologies} 
Ensuring data privacy and security when applying AI to the industry is indeed important. Training LLMs requires vast amounts of data, and storing data centrally poses significant risks, making it easier for hackers to attack and extract user information. To address this issue, federated learning~\cite{nguyen2024efficient, jiang_feibo2024wireless_application_arxiv} is a promising approach that maintains data privacy by keeping data on individual devices rather than transferring it to a central server. In federated learning, the model is trained locally on each device using the device's data, and only the model updates are shared with the central server. This prevents sensitive data from leaving the device, thus protecting user privacy. Furthermore, methods that avoid centralized data storage can reduce the risk of single-point attacks by attackers. Technologies such as secure multi-party computation (SMPC)~\cite{goldreich1998secure} and homomorphic encryption~\cite{acar2018survey} can be integrated with federated learning to enhance privacy and security. Specifically, SMPC allows multiple parties to jointly compute functions of their inputs while maintaining the privacy of those inputs. Meanwhile, homomorphic encryption enables computations to be performed directly on encrypted data without decryption, ensuring the security of the entire data chain.

In addition, split learning~\cite{thapa_chandra2022split_learning} is another powerful approach that can bolster data privacy and security. By distributing the training process between edge devices and a central server, split learning reduces the risk of data leakage. In this method, only the intermediate layers of the model are shared with the server, while the raw data remains on the edge devices. This approach significantly minimizes the chances of sensitive information being intercepted or decrypted through simple extraction techniques. An overview of potential solutions is illustrated in Fig.~\ref{fig:llm_for_wireless_solutions}.

\section{Wireless Networks for RL-based LLMs}
\label{sec:empower_llms_with_wireless}

\begin{wrapfigure}{r}{0.5\textwidth}
    \centering
    \includegraphics[width=0.95\linewidth]{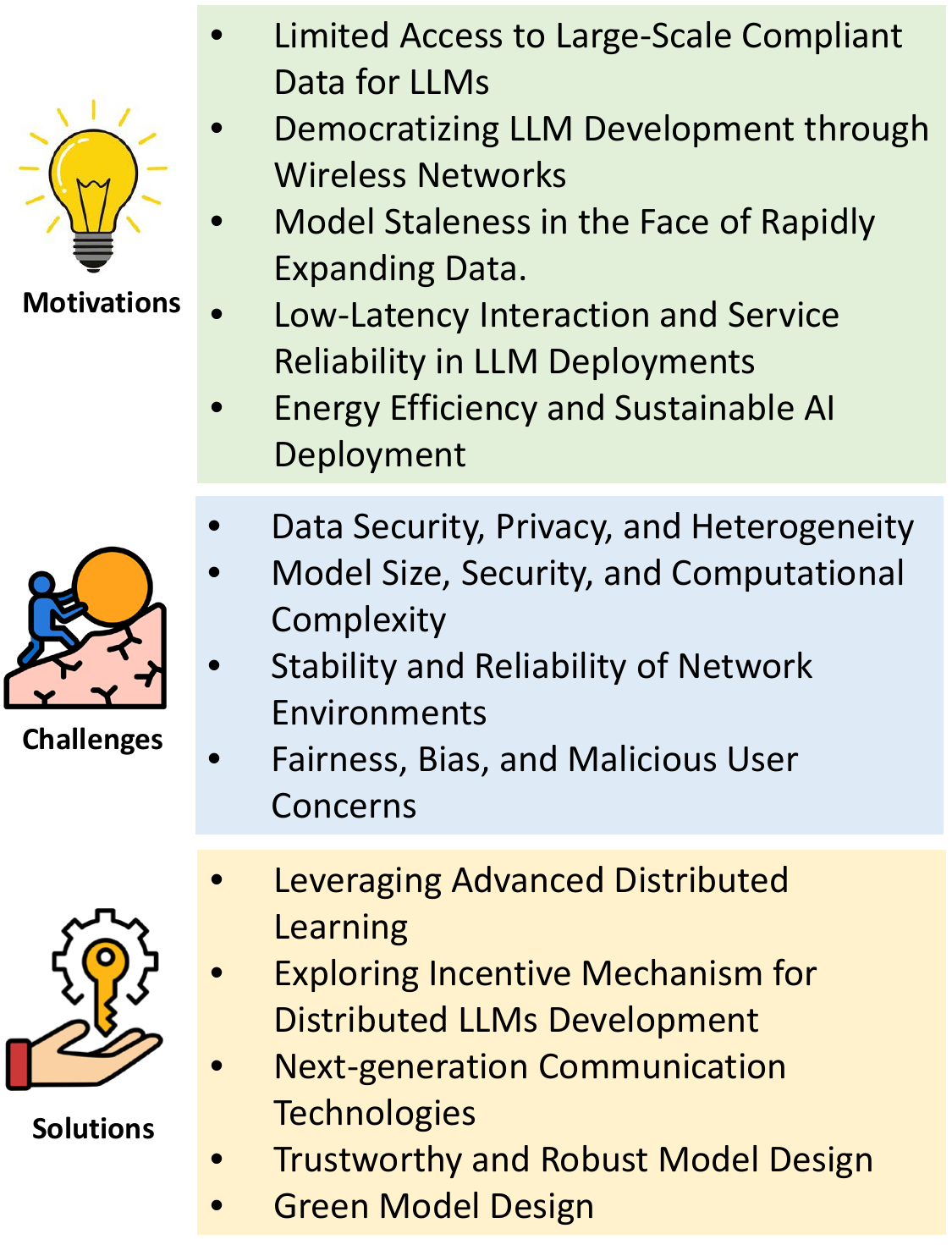}
    \caption{Motivations, challenges, and potential solutions for empowering LLMs with wireless communication systems.}
    \label{fig:wireless_for_llm}
\end{wrapfigure}

In this section, we outline the key motivations for leveraging wireless communication systems in reinforcement learning (RL)-based large language models (LLMs), including the limited access to large-scale compliant data, the democratization of LLM development through wireless networks, and the challenge of model staleness amid the rapid expansion of data. We then explore the challenges of integrating wireless communication systems into RL-based LLMs from the perspectives of data, models, environment, and users. Finally, we discuss potential solutions to address the challenges mentioned above. Fig.~\ref{fig:wireless_for_llm} offers a comprehensive overview of the motivations, challenges, and potential methods for integrating wireless communication systems into LLMs.

\subsection{Motivations for Empowering RL-based LLMs with Wireless Networks}
\label{sec:motivation_empower_llms_with_wireless}

\subsubsection{Limited Access to Large-Scale Compliant Data for LLMs} Given a large number of parameters in LLMs, training them on vast amounts of high-quality data from sources like the Internet, book corpus, Wikipedia, and other large-scale datasets can be prohibitively expensive~\cite{zhao2024generative,zhang2024role}. For instance, training a model like GPT-3, which has 175 billion parameters, requires substantial computational resources and access to massive datasets, often reaching millions of dollars in costs for infrastructure and energy~\cite{Wallstreetcn2024}. 
However, access to such datasets is often restricted to privacy regulations, such as GDPR~\cite{voigt2017eu} or HIPAA~\cite{cohen2018hipaa}, posing significant challenges, as many publicly available data sources may not meet these legal and ethical requirements. For example, in healthcare applications, patient records are protected by strict privacy laws, making it difficult to train medical LLMs with diverse, real-world data.

Wireless communication systems, with their distributed nature and continuously advancing hardware and software capabilities, offer a promising solution for achieving decentralization and joint data aggregation across distributed sources. By leveraging technologies like continual learning~\cite{wang2024comprehensive}, federated learning~\cite{qiao2023knowledge}, and secure multi-party computation~\cite{zhou2024secure}, model training can be distributed across various edge devices. Each edge device can collaborate to update the model based on its private dataset, eliminating the need for a central organizer to access the original data, thus ensuring privacy-preserving large-scale distributed training. Additionally, new edge devices can join the network at any time, seamlessly providing additional resources to enhance the scalability of model training. For example, in autonomous driving, vehicles can collaboratively train LLMs by sharing model updates over wireless networks without exposing sensitive sensor data. This method not only ensures compliance with data privacy regulations but also enhances the model’s adaptability to real-world conditions.

\subsubsection{Democratizing LLM Development through Wireless Networks} It is widely recognized that the investment required for AI research and development is substantial, especially for companies developing large models~\cite{mou2019artificial}. For example, Sequoia's research indicates that AI investment will exceed revenue by over \$120 billion in 2024, potentially rising to \$500 billion in 2025~\cite{widder2024watching}. Microsoft's CFO has mentioned that the company may need to wait 15 years or more to recover the tens of billions of dollars invested in LLMs~\cite{cohan2024generative}. This underscores the enormous cost involved in training and maintaining such models. Currently, the development and deployment of state-of-the-art LLMs are monopolized by a few well-funded tech companies, limiting the participation of other organizations and stifling technological innovation and democratization.

However, it is increasingly feasible for wireless communication systems, particularly in federated or decentralized environments, to democratize access to resources and data. This shift is expected to enable smaller, capital-driven companies or individual entities to contribute to and drive the development of LLMs. Meanwhile, it is worthwhile to highlight that the vision of democratizing LLM development is becoming increasingly achievable. For instance, \cite{douillard2023diloco} proposes a method called DiLoCo for training LLMs in a distributed environment. This approach to collaborative model development and maintenance reduces the technical barriers created by resource concentration and promotes innovation. It also minimizes reliance on large-scale centralized infrastructure, allowing LLM development, maintenance, and access to be more broadly distributed and democratized.

\subsubsection{Model Staleness in the Face of Rapidly Expanding Data} Model staleness arises when pre-trained models are trained on outdated datasets and fail to keep pace with the rapid expansion of new data~\cite{chen2022sapipe}. As fresh information continuously emerges, models relying on outdated training data and parameters struggle to adapt to evolving real-world scenarios. This challenge is particularly pronounced in wireless environments, where conditions are highly dynamic and unpredictable. For instance, in autonomous driving, sensor data collected by vehicles changes constantly due to varying road conditions, weather, and traffic patterns~\cite{zhang2023perception}. In such cases, pre-trained models that lack real-time adaptation capabilities may become ineffective in responding to new and unforeseen situations, potentially leading to hazardous incidents or accidents.

Wireless communication systems offer a promising solution to this problem through edge computing~\cite{qiao2023prototype}. By leveraging edge computing techniques such as on-device computing~\cite{pandey2020crowdsourcing} and computation offloading~\cite{hossain2023computation}, LLMs can be incrementally fine-tuned with new data, enabling continuous adaptation without requiring retraining from scratch. Moreover, reinforcement learning techniques such as RLHF~\cite{ouyang_long2022RLHF} and RLAIF~\cite{bai_yuntao2022RLAIF}, when integrated into wireless networks, allow models to autonomously update and refine their performance over time based on real-world interactions. This ensures that LLMs stay up to date with the latest data trends, enhancing their performance and reliability in dynamic environments.

\subsubsection{Low-Latency Interaction and Service Reliability in LLM Deployments}
Traditional LLMs are typically deployed in the cloud and hosted on central servers, requiring users to access them via APIs~\cite{Wallstreetcn2024}. This setup often introduces delays, particularly when bandwidth is limited or the model is large. In addition, cloud servers are vulnerable to outages and cyberattacks, as demonstrated by the recent DDoS attack on DeepSeek~\cite{xu2025dark}, which temporarily disrupted user access. In such cases, users who rely heavily on centralized servers may experience high latency or service interruptions, significantly degrading the overall user experience.

Deploying LLMs on edge devices within wireless networks presents a promising solution to these challenges by enabling faster inference and reducing vulnerability to centralized failures. Moreover, edge devices can collaborate to distribute model inference tasks, bringing computation closer to the data source. This allows real-time data processing and analysis with minimal latency, ensuring swift adaptation to changing conditions. For instance, in smart cities, traffic control systems can leverage edge devices to process real-time traffic data and dynamically update LLMs for decision-making. This decentralized approach reduces dependence on cloud infrastructure, enhances service speed and reliability, and enables more responsive and adaptive AI-driven applications.

\subsubsection{Energy Efficiency and Sustainable AI Deployment}
The training and inference of LLMs demand substantial computational resources, leading to significant energy consumption and an increased carbon footprint~\cite{iftikhar2024reducing}. Training these large-scale models often requires running thousands of GPUs or TPUs for extended periods, consuming vast amounts of electricity and necessitating extensive cooling infrastructure to maintain hardware stability. For instance, the latest Grok-3~\cite{reyes2024evaluating}, released in 2025, required a cumulative training time of 200 million hours on 200,000 NVIDIA H100 GPUs. This immense energy demand not only raises concerns about the sustainability of AI development but also poses challenges for deploying such models in resource-constrained environments. In remote or low-resource regions, the limited power supply may make it impractical to train or run LLMs on traditional infrastructure.

Wireless communication systems, particularly when integrated with techniques like federated learning~\cite{qiao2023boosting}, offer a promising solution by leveraging distributed optimization algorithms to reduce communication overhead and energy consumption. For example, the DiLoCo~\cite{douillard2023diloco} algorithm enables LLM training on a network of devices with poor connectivity, significantly lowering communication frequency and power usage by performing multiple local optimization steps before transmitting updates. Building on this, Photon~\cite{sani2024photon} further enhances collaborative training under weak connectivity and low-bandwidth conditions, introducing the first cost-effective, global-scale LLM pre-training system that achieves twice the convergence speed of previous approaches like DiLoCo. By harnessing wireless communication to facilitate decentralized LLM training, this approach not only enhances energy efficiency but also supports the sustainable deployment of AI, mitigating environmental impact while ensuring broader accessibility.

\subsection{Challenges}
\label{sec:challenges_empower_llms_with_wireless}

\begin{figure*}[t]
    \centering
    \includegraphics[width=\linewidth]{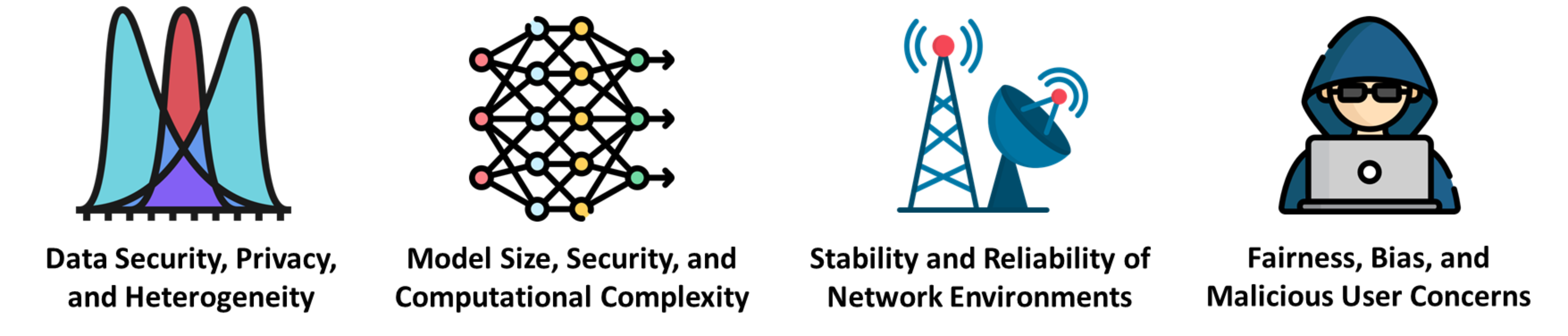}
    \caption{Challenges for empowering LLMs with wireless communication systems.}
    \label{fig:wireless_for_llm_challenges}
\end{figure*}

Integrating wireless communication systems into RL-based LLMs presents several challenges, which can be broadly categorized into four key aspects: data, model, environment, and user, as illustrated in Fig.~\ref{fig:wireless_for_llm_challenges}.

\subsubsection{Data Security, Privacy, and Heterogeneity} 
To date, all of the studies on RL-based LLMs are trained on a centralized server with a massive computing capacity, which demands data collection across a large number of devices and domains to train the model successfully~\cite{ren2024industrial}. However, the data are usually generated at the device sites and often exhibit characteristics of heterogeneity, including domain heterogeneity~\cite{qiao2024fedccl}, which is a newly summarized concept for this type of variability. These data need to be transmitted to the server through the wired or wireless environment. As a result, the transmission through wireless channels poses a significant risk of eavesdropping and adversarial attacks~\cite{qiao2024logit}, which both create severe situations for data transmission for LLMs. In the first case, an eavesdropper can process the passive sniffing by secretly capturing the signal without any interaction with the network or setting up similar configurations with the legitimate receiver to get the data. Then, the signal can be decoded and used for illegitimate purposes. Additionally, the eavesdropper can further interfere with the receiver's signal reception, causing it to deviate from its original meaning, which can degrade the data quality and destroy the training process of LLMs. Presumably, we can successfully obtain the data without any adversarial attack or privacy issue, but there will still be problems with the heterogeneity distribution of data. The heterogeneous data distribution indicates the unbalance among different types of data. For example, the data in the medical field is extremely scarce, while the data related to sensor networks in smart cities is redundant~\cite{liu2024survey_medical}.

\subsubsection{Model Size, Security, and Computational Complexity.} As discussed in the motivation section, by leveraging the distribution of devices across physical space, LLMs can be sent down and trained on local devices, thereby reducing the potential risks of transmission of the training data over the wireless network. Nevertheless, the approach consists of several significant drawbacks, such as the large size of the LLMs model (bandwidth consumption), the vulnerability to attacks, and the computing capacity of local devices~\cite{hasan2024distributed}. In distributed learning, the model is transmitted instead of the data, and this approach usually lowers the communication cost; however, the size of the LLMs is much larger than the normal cases, and thus, it still encounters difficulty when the communication resource is restrained~\cite{ren2024industrial}. Another disadvantage of distributing the LLMs to local devices is the malicious training agents, whose target is to degrade the performance of LLMs by attributing their useless weights and updates to the global model aggregation at the central server~\cite{chen2022sapipe}. Lastly, the dynamic computing capability of local devices should be considered; specifically, this can maximize the utilization of power devices and go easy on the low-computing ones. In real-world scenarios, the approach leads to sustainable and cost-effective deployments by putting more workload and trust in reliable and high-computing agents, while the unimportant tasks should be trained by low-computing agents.

\subsubsection{Stability and Reliability of Network Environments} In addition to the challenges related to data and models in wireless communications, the characteristics of the network environment itself can significantly impact the effectiveness of training or deploying LLMs. The quality of data and model parameters may be compromised when transmitted over noisy wireless channels, leading to performance degradation or even loss of critical information due to communication congestion~\cite{javaid2021optimized}. Therefore, ensuring the robustness and efficiency of LLMs in dynamic wireless environments is essential yet remains a significant challenge. The property of the wireless environment that creates difficult problems is the swift transformations in the physical surroundings, for example, the moving object blocks the line-of-sight between the transmitter and receiver in a short time without the acknowledgment of both devices~\cite{bensky2019short}. This scenario is more likely to occur when the current wireless network prioritizes fast transmission and supports a large number of connected devices while reducing the correction protocol among them to minimize communication and computation latency. When transmitting RL-based LLMs with large sizes, the primary goal is to ensure a stable and reliable network. Otherwise, varying levels of channel noise can affect different layers in different ways, making it impossible to eliminate noise across the entire model and achieve optimal performance. Additionally, the unstable connections and fluctuating bandwidth in wireless networks can disrupt the RL training loop, resulting in incomplete learning cycles and diminished model performance~\cite{chen2020reinforcement}. Therefore, maintaining consistent communication between the agent and the server is vital to ensure that RL-based LLMs can adapt and learn effectively in real-world applications.




\subsubsection{Fairness, Bias, and Malicious User Concerns} Due to the decentralized and heterogeneous nature of wireless networks, empowering LLMs through these networks presents significant challenges in fairness, bias, and security. First, users in a wireless network have varying levels of connectivity and resources, which can lead to fairness issues~\cite{wu2025fairness}. Those with stronger connections and better resources may dominate the learning process, while users with weaker network conditions risk being underrepresented, resulting in imbalanced model optimization. Second, bias can arise if the majority of devices in the network are concentrated in specific regions, leading the model to favor certain characteristics while neglecting others~\cite{javed2025robustness}. For example, if the model is predominantly trained on English data, it may struggle with resource-poor languages, generating less fluent or even inaccurate translations. Similarly, if historical hiring data exhibits a preference for male candidates in technical roles, the model may disproportionately recommend such positions to men, even when female candidates have equivalent qualifications, thereby reinforcing gender biases in technical fields~\cite{mujtaba2024fairness}. Finally, since RL-based LLMs rely on continuous interaction with users and the environment, they are vulnerable to malicious manipulation~\cite{hurst_aaron2024gpt4o}. Malicious users may inject poisoned data, manipulate feedback signals, or launch adversarial attacks~\cite{qiao2024fedccl}, compromising model integrity and security. Addressing these challenges necessitates the implementation of fairness-aware learning mechanisms, bias mitigation strategies, and robust adversarial defenses to ensure that RL-based LLM frameworks in wireless networks are fair, reliable, and secure.

\subsection{Potential Solutions}
\label{sec:solutions_empower_llms_with_wireless}

\begin{figure*}[t]
    \centering
    \includegraphics[width=\linewidth]{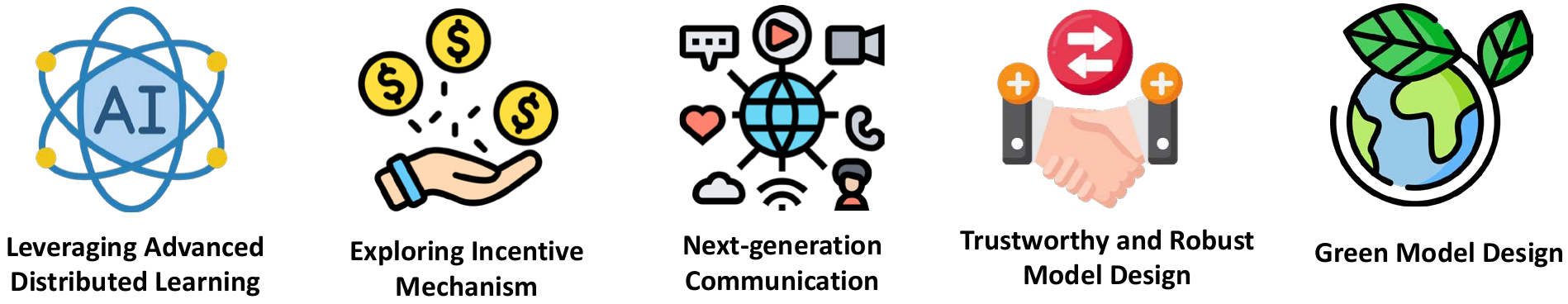}
    \caption{Potential solutions for empowering LLMs with wireless communication systems.}
    \label{fig:wireless_for_llm_solutions}
\end{figure*}

\subsubsection{Leveraging Advanced Distributed Learning} To enhance LLMs in wireless communication systems, advanced distributed learning techniques can be employed to improve model performance and adaptability. Traditional LLMs rely on large-scale centralized data for training, necessitating compliance with data protection and privacy regulations, such as GDPR~\cite{voigt2017eu}. Distributed training over wireless networks, such as federated learning~\cite{mcmahan2017communication}, enables LLMs to be trained across numerous local computing clusters, thereby avoiding direct, centralized data storage and enhancing privacy protection capabilities. However, a risk of data leakage persists when distributed local devices collaborate to train models. For instance, during the model update process, participating devices may inadvertently reveal the statistical characteristics of their local data or model parameters, allowing attackers to infer sensitive information about certain users. To mitigate this risk, techniques such as differential privacy~\cite{abadi2016deep} can be implemented to ensure that individual data contributions remain confidential, while secure multi-party computation~\cite{zhou2024secure} can aggregate model updates without exposing any raw data. Additionally, secure aggregation protocols can help maintain data confidentiality during training.

Moreover, local data sources may exhibit heterogeneity, which can negatively impact the model's generalization ability. To address this issue, prototype-based~\cite{le2025mitigating,qiao2024fedccl} domain alignment techniques can align the model with the global data distribution. Furthermore, generative artificial intelligence technologies, such as diffusion models~\cite{croitoru2023diffusion,pmlr-v37-sohl-dickstein15,10049010,ho2020denoising} and GANs~\cite{xia2022gan,goodfellow2014generative,mao2017least}, offer promising solutions by generating synthetic samples that simulate various data distributions, thus alleviating the effects of data heterogeneity and enhancing the model's robustness and generalization capability.

\subsubsection{Exploring Incentive Mechanism for Distributed LLMs Development} Exploring incentive mechanisms that promote the development of distributed LLMs is critical for the advancement of the AI community, as it can effectively increase participant enthusiasm and encourage more devices and users to contribute their computing resources and data~\cite{tran2025multi}. Consequently, the design of these incentive mechanisms must be both effective and fair, taking into account the differences in computing resources and power across various regions where participants are located. In wireless networks, devices may be reluctant to participate in training or share data due to concerns about privacy, potential misuse of information, or simply a lack of perceived benefits~\cite{chen2024big}. Therefore, future technical designs should provide an open and transparent framework for the use of data and computing resources while also offering appropriate rewards to users who contribute, such as free access to LLM API resources. By establishing a balanced incentive structure, we can foster a collaborative environment that promotes the development of distributed LLMs while ensuring user participation and data protection.

\subsubsection{Next-generation Communication Technologies} The development of LLMs enabled by wireless network systems presents challenges such as network bandwidth limitations, which can hinder model training and updates. Therefore, next-generation communication technologies such as 6G~\cite{jiang2021road}, semantic communication~\cite{yang2022semantic}, and integrated sensing and communication (ISAC)~\cite{adhikary2023intelligent,adhikary2022edge} offer promising solutions. 6G technology is designed to provide ultra-reliable low-latency communication and significantly increased data rates, facilitating the seamless transmission of LLMs updates and real-time interactions. Semantic communication improves efficiency by focusing on the transmission of meaningful information rather than raw data, thereby significantly reducing bandwidth requirements and latency. This is especially beneficial for LLMs, as it enables faster and more efficient data exchange. Additionally, ISAC integrates communication with environmental sensing, allowing for optimized resource utilization and enhanced situational awareness. By leveraging these advanced communication technologies, we can establish a more robust framework for distributed LLMs, ensuring effective model performance while addressing critical issues related to data privacy and computational demands.

\begin{figure*}[t]
    \centering
    \includegraphics[width=0.70\linewidth]{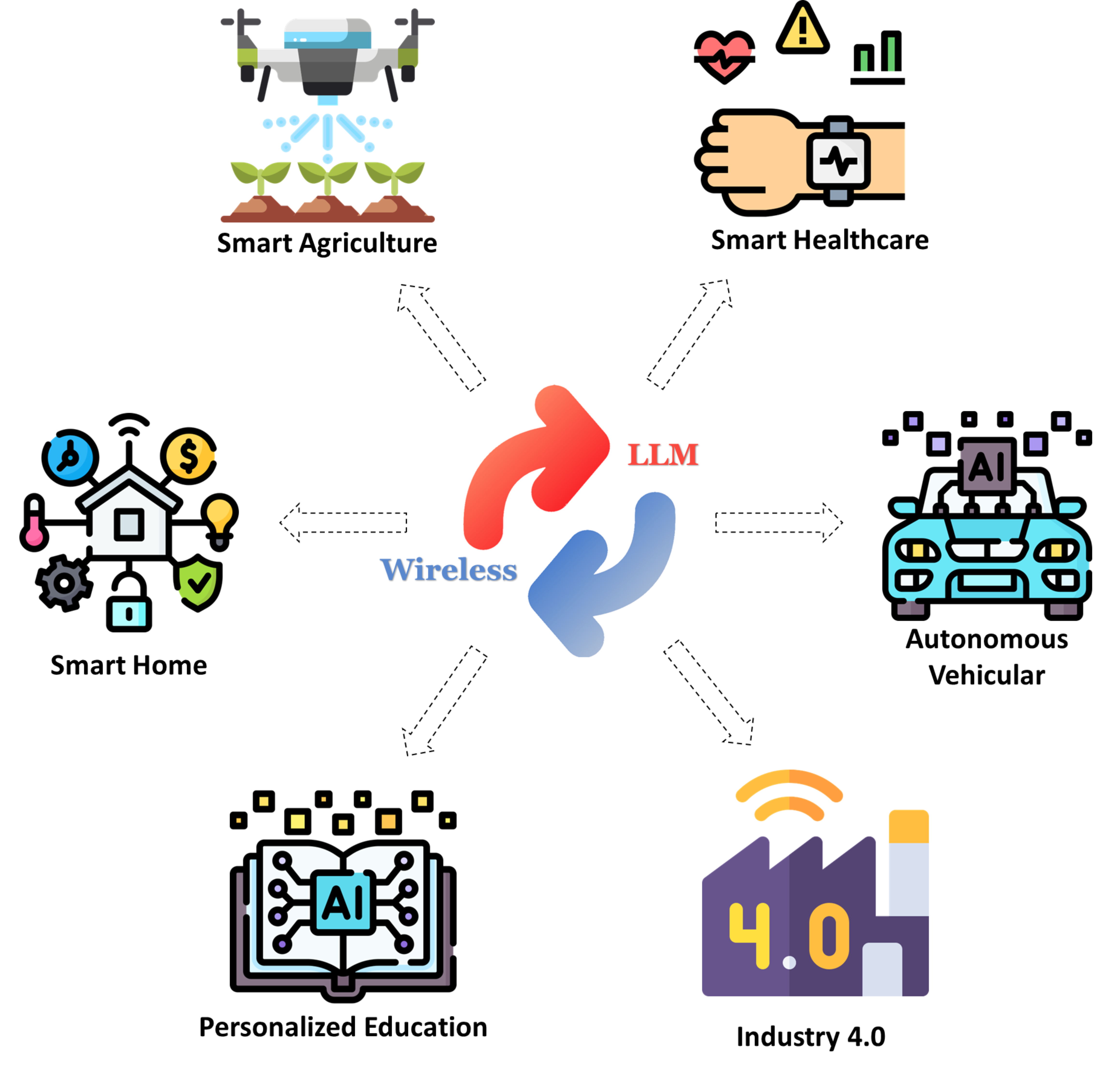}
    \caption{Illustration of applications demonstrating the mutual empowerment between wireless communication systems and LLMs.}
    \label{fig:applications}
\end{figure*}

\subsubsection{Trustworthy and Robust Model Design} In order to effectively enable LLMs over wireless networks, it is essential to design trustworthy and robust models that address key challenges such as unfairness, bias, and unreliable environments. Trustworthy AI~\cite{li2023trustworthy} emphasizes the reliability, transparency, and ethical considerations of AI systems, ensuring that users can trust the decisions made by these systems. Training models based on data from a specific region can lead to biased decisions, reinforcing stereotypes that associate engineers with men and nurses with women. While distributed model training can help mitigate this risk, the black-box nature of model decisions often leaves users unaware of the underlying processes. This lack of understanding can foster doubts about the model's fairness, leading to a lack of trust in its outputs and potentially exacerbating bias and discrimination against certain groups. Therefore, the development of these models should incorporate several core elements: transparency, which enables users to understand the reasoning behind AI decisions; fairness, which seeks to minimize bias and discrimination; privacy protection, ensuring user data remains secure; and traceability, allowing the decision-making process to be tracked.

In addition, robust AI~\cite{tocchetti2022ai} focuses on the stability and adaptability of AI systems under various conditions, particularly in the face of uncertainty or interference. This encompasses maintaining performance despite disturbances, adapting to environmental changes and new data distributions, and demonstrating fault tolerance to manage errors or anomalies without significant failures. For instance, to mitigate the impact of adversarial attacks~\cite{qiao2025towards}, adversarial samples can be intentionally introduced during the training phase to reduce the likelihood of successful attacks. By creating models that users can trust, that perform well across diverse environments, and that adhere to ethical standards, we can enhance user trust and participation in LLM applications over wireless networks.

\subsubsection{Green Model Design} The development of LLMs demands substantial hardware and energy resources, significantly increasing their carbon footprint. Therefore, designing green models is essential for building sustainable AI systems. To minimize environmental impact, strategies such as quantization~\cite{wenqu_shao2024ICLR}, model sharding~\cite{quan2024recent}, knowledge distillation~\cite{qiao2025towards}, and efficient architectural designs can be employed. Quantization reduces the storage and computational requirements of a model by lowering the bit precision of weights and activations, thereby decreasing energy consumption and carbon emissions. Model sharding partitions large models into smaller, more manageable segments, enabling distributed devices to train different model components collaboratively. This not only alleviates the computational burden on individual devices but also enhances training efficiency by allowing each device to focus on a specific shard. Knowledge distillation further optimizes resource usage by transferring knowledge from a larger, complex teacher model to a smaller, lightweight student model, ensuring efficient inference with reduced computational costs.

In addition, efficient architectural design plays a critical role in reducing carbon emissions. For instance, in long-text modeling, traditional full attention mechanisms exhibit quadratic complexity with respect to sequence length, leading to high memory and computation costs. Recent advancements in native sparse attention~\cite{Yuan2025native} have optimized conventional full attention, reducing memory access to just 1/11.6 of its original requirement without compromising performance. By integrating these technologies into wireless network systems, we can advance LLM development while ensuring a more energy-efficient and sustainable approach, allowing these models to operate effectively in diverse environments with minimal environmental impact. An overview of potential solutions is illustrated in Fig.~\ref{fig:wireless_for_llm_solutions}.

\section{Applications}
\label{sec:applications}

In this section, we explore four applications as illustrated in Fig.~\ref{fig:applications}, including intelligent ecosystems, autonomous vehicular networks, smart manufacturing and Industry 4.0, and personalized education. These examples highlight the mutual empowerment of wireless networking systems and LLMs. While multiple other applications may exist, these four serve as representative cases to facilitate discussion.

\subsection{Intelligent Ecosystems}
The integration of wireless network systems with LLMs is revolutionizing smart ecosystems across various domains. In smart cities~\cite{kalyuzhnaya2025llm}, LLMs facilitate real-time data processing from IoT sensors to optimize traffic management, energy distribution, and public safety, while wireless networks ensure seamless communication between interconnected devices. This continuous stream of data can further enhance LLM performance. Similarly, in smart healthcare~\cite{lai2022healthcare}, AI-driven diagnostics and automated patient monitoring leverage wireless connectivity to enable remote medical services and real-time consultations. In smart homes~\cite{rivkin2024aiot}, LLM-powered virtual assistants, such as Apple Home and LLM Siri, can improve voice recognition, personalized automation, and security system integration, creating a more intuitive living experience. Moreover, in smart agriculture~\cite{li2024foundation}, the synergy between LLM-driven analytics and wireless sensor networks fosters precision agriculture by optimizing irrigation, pest control, and crop yield predictions, ultimately boosting productivity. This mutual empowerment also extends to wearable technology, which can provide real-time language translation, biometric monitoring, and personalized user experiences, with wireless connectivity ensuring seamless data exchange and continuous updates. By combining the intelligence of LLMs with the adaptability of wireless infrastructure, these smart ecosystems would become more autonomous, responsive, and mission-critical, driving advancements in automation, decision-making, and operational efficiency across diverse applications.

\subsection{Autonomous Vehicular Networks}
Developing an autonomous vehicular network involves a multifaceted development that encompasses numerous challenges. Advanced sensing technologies, such as LIDAR, radar, and cameras, are crucial for accurate perception and decision-making, requiring sophisticated AI algorithms for effective operation~\cite{thakur2024depth}. Real-time response is essential for the effective operation of these systems. Integrating LLMs into wireless communication can mitigate the burden of processing data by employing semantic communication and contextual understanding. Moreover, LLMs play a vital role in optimizing network traffic in both wireless communication and vehicular routing~\cite{hu2025llm}. By selecting the most efficient routes for data transmission, LLMs can enhance real-time response capabilities while providing valuable information for AI to control the vehicle.

For example, LLMs can analyze and interpret complex data from various sensors and communication channels, enabling smarter decision-making and enhanced situational awareness~\cite{hu2025llm}. Additionally, wireless networks facilitate dynamic communication between vehicles and infrastructure, providing real-time updates on traffic conditions, potential hazards, and optimal routes. This synergy is expected to lead to a safer and more efficient network of autonomous vehicles that can adapt to changing environments and user needs. Recent research introduces DriveGPT4~\cite{xu2024drivegpt4}, an interpretable end-to-end autonomous driving system that extends LLMs into the multimodal domain to process road scene data. DriveGPT4 has demonstrated promising performance in predicting low-level vehicle control signals, including steering, acceleration, braking, and gear shifting.

\subsection{Industry 4.0}
Digital twins have become essential in smart manufacturing and Industry 4.0, changing how industries work~\cite{nvidia_digital_twin}. One new trend is using real-time AI for industrial automation, which involves creating an AI gym where robots and AI agents can be trained and tested in complex environments such as storage facilities~\cite{nvidia_digital_twin}. This requires many techniques, including LLMs, for control, decision-making, sensing, and communication, as well as wireless connections for smooth interactions between agents and servers.

For example, in storage management, an AI robot needs to load and organize products in specific locations~\cite{murphy2019introduction}. The path from loading to unloading has many variables that can affect the efficiency of this action. To handle this, the AI agent collects data from sensors, cameras, microphones, and other sources, sending this information to the server for processing. Any delays in sending data or commands can cause financial losses and other problems. Therefore, it's crucial to have a strong system that can make quick and accurate decisions for these applications using LLM and wireless networks.


\subsection{Personalized Education}
Personalized education considers individual learning styles, preferences, and pace to enhance student engagement, foster a deeper interest in learning, and ultimately improve academic performance~\cite{das2025llm_security}. The integration of wireless networks and LLMs has the potential to revolutionize personalized education by providing tailored learning experiences for students. Through real-time data transmission from devices, educators can adjust teaching materials based on the progress of different students, addressing diverse learning needs. For example, personalized learning platforms can deliver content suited to various learning styles, with visual learners receiving more graphics and video materials, while auditory learners engage with audio lectures~\cite{zhang2024enhancing}. Teachers can analyze students' learning data in real-time to identify those needing additional support and provide targeted guidance. In addition, adaptive learning systems can dynamically modify learning paths according to students' performance, ensuring they learn at an appropriate difficulty level. Furthermore, educational institutions can develop commercial personalized education tools to optimize curriculum settings and resource allocation, promoting the widespread adoption and advancement of personalized education. By leveraging these strategies, the combination of wireless networks and LLM would significantly enhance the effectiveness of personalized education, allowing every student to benefit from a customized learning experience.

\section{Future Directions}
\label{sec:directions}
In this section, we present a vision for the future directions of the development of LLMs for wireless networks, including quantum LLMs, on-device LLMs, neural-symbolic LLMs, and embodied AI agents.

\subsection{Quantum LLMs}
Quantum computing~\cite{gyongyosi2019survey} is a new computing paradigm that utilizes the principles of quantum mechanics to manipulate quantum information units for computation. By employing quantum bits (qubits) to process information in parallel, it achieves greater computational power and the capacity to solve more complex problems than traditional computing methods~\cite{gyongyosi2019survey}. The advancement of quantum computing is anticipated to have a profound impact on existing technologies, including LLMs. For instance, quantum computing has the potential to significantly improve the efficiency of computing, improve data security, and speed up model training, thereby greatly boosting the performance of LLMs~\cite{behura2025challenges}. However, to fully realize the benefits of quantum computing, it is essential to transform conventional computer algorithms into quantum algorithms specifically designed for quantum architectures. Research on quantum-enhanced LLMs needs to prioritize optimizing model architecture design, model training and inference, and resource allocation to maximize the advantages provided by quantum computing~\cite{behura2025challenges}. In addition, exploring quantum-inspired neural networks and quantum machine learning frameworks could unlock new paradigms in language understanding, reasoning, and data processing.

\subsection{On-Device LLMs}
LLMs must evolve beyond simple response generation to deeply integrate into machine learning (ML) workflows while operating efficiently on edge devices~\cite{zhang2024llm_edge_computing}. They should assist with complex ML-specific tasks such as model selection, hyperparameter tuning, feature engineering, and AutoML while leveraging multimodal capabilities to process diverse data types, including text, images, and structured datasets~\cite{zhao2019edge}. However, deploying such advanced LLMs on-device introduces challenges related to model compression, energy efficiency, and latency. Reinforcement learning (RL) offers a solution by enabling adaptive execution under hardware constraints, dynamically optimizing computational cost and accuracy~\cite{chen2020reinforcement}. Federated reinforcement learning~\cite{qi2021federated} further enhances personalization while maintaining privacy, allowing LLMs to learn from user interactions without cloud dependence. RL can also optimize resource allocation, quantization, and pruning to ensure energy-efficient performance while enabling real-time adaptation to environmental feedback~\cite{10306287}. This makes RL-based LLMs ideal for real-time applications such as autonomous agents and interactive assistants. By integrating RL with on-device AI, LLMs can become more autonomous, efficient, and privacy-preserving, leading to a new generation of AI systems that seamlessly support ML tasks while operating independently on edge devices.


\subsection{Neural-Symbolic LLMs}
Current research on LLMs primarily focuses on natural language processing~\cite{kumar2024large,yan2024protecting,das2025llm_security}. However, world knowledge extends beyond natural language definitions and expressions. For instance, data in wireless environments often takes non-natural language forms, such as XML and JSON, which are used for data exchange and information structure descriptions to facilitate communication between different systems. In addition, CMOS sensors in mobile phones typically interact with the main control chip through communication interfaces like I2C or SPI, utilizing specific command sets~\cite{malviya2020tiny}. Therefore, future developments in LLMs should prioritize optimizing models for symbolic languages, enabling them to effectively process and understand various data formats and communication protocols, thereby broadening their applicability across diverse domains. Furthermore, since symbolic languages are implemented through structured rules and representations, they offer clear traceability, logical reasoning capabilities, and enhanced interpretability~\cite{pulicharla2025neurosymbolic}. Combining symbolic language with LLMs, based on traditional natural language training, can make the reasoning process transparent and easy to understand while accurately expressing complex concepts and relationships.

\subsection{Embodied AI Agents}

Embodied AI agents are intelligent systems built around LLMs as core controllers, enabling them to independently perform tasks, make decisions, and adapt to their environments through reasoning and learning~\cite{zhang_ruichen2024wireless_application}. This vision is realized through models such as AutoGPT~\cite{yang2023auto} and BabyAGI~\cite{bieger2015safe}, which possess strong general problem-solving capabilities and can function as the brains of autonomous AI agents. However, to fully realize the potential of autonomous AI agents, the evolution from passive models to those capable of reasoning, decision-making, environmental perception, dynamic learning, and continuous updating is essential. For instance, RL techniques can empower these agents to learn from feedback, enhance their responses, and optimize decisions dynamically~\cite{wang2022policy_arxiv}. In other words, unlike traditional LLMs, RL-enhanced AI agents can plan tasks, solve multi-step problems, and adapt to changing circumstances. By integrating multimodal capabilities, these agents can process text, visual information, and actions, making them suitable for applications such as autonomous research assistants, adaptive tutoring systems, and robotic automation. Furthermore, on-device RL can enhance efficiency by reducing reliance on cloud resources and improving privacy~\cite{pandey2020crowdsourcing,9759480}. Future research should focus on developing scalable RL algorithms, hierarchical learning for task planning, and improved exploration strategies to create truly self-learning AI agents that can operate autonomously.

\section{Societal Impact}
\label{sec:impact}
Although the development of wireless networks can be significantly empowered by advancements in AI, particularly in LLMs, and AI is reshaping wireless technology, the resulting societal impacts must be considered.

\subsection{Ethics and Morality} The advancement of AI and wireless network technologies raises various ethical and moral issues. First, many traditional jobs may be replaced by AI systems that can perform tasks more efficiently and at a lower cost, leading to significant unemployment and exacerbating socioeconomic inequality~\cite{kumar2024large}. Second, unequal access to AI technology may widen the gap in wealth and opportunities~\cite{farahani2024artificial}. Those who can leverage AI may gain substantial economic benefits, while those without access risk marginalization, resulting in social stratification and unfairness. Third, AI lacks human emotions and empathy, which can lead to decisions misaligned with human values, especially in complex interpersonal and moral situations~\cite{parthasarathy2024ultimate}. Fourth, AI systems may make mistakes and produce illogical or harmful outcomes, potentially undermining the rights of individuals with diminished autonomy due to age or illness~\cite{das2025llm_security}. Fifth, AI systems designed to be overly intelligent may create unforeseen and harmful consequences~\cite{tocchetti2022ai}. Sixth, as AI evolves to achieve self-improvement and surpass human intelligence, unpredictable outcomes could emerge, resulting in a loss of human control and significant ethical and social challenges~\cite{parthasarathy2024ultimate}. Finally, as AI becomes more intelligent and autonomous, discussions about whether it should possess rights are increasing, prompting deep reflections on the ethical and legal implications of balancing rights and responsibilities between humans and AI~\cite{tocchetti2022ai}.

\subsection{Transparency and Accountability} AI systems that leverage the mutual enhancement between wireless networks and AI technologies necessitate a high level of transparency and accountability~\cite{quttainah2024cost}. Proof-of-concept projects related to AI often stall during testing due to a lack of trust in the results generated by AI models~\cite{afroogh2024trust}. This distrust arises from several factors, including the opacity of AI decision-making processes and the potential for inherent biases in algorithms. When stakeholders do not clearly understand how decisions are made, they may question the reliability and fairness of AI outputs, leading to hesitant adoption. To address these issues, it is crucial to establish transparent processes that clarify how AI systems operate and make decisions. This can involve providing comprehensive documentation, clear explanations of algorithmic logic, and strategies to identify and mitigate bias. In addition, accountability measures should be implemented to ensure that AI systems comply with ethical standards and are held responsible for their decisions~\cite{farahani2024artificial}. By fostering transparency and accountability, we can enhance trust in AI technology and facilitate its successful integration into wireless networks and broader societal applications.

\subsection{Legal Considerations} The integration of AI with wireless networks presents significant legal challenges. When fully AI-driven systems make erroneous or harmful decisions that violate legal standards, determining liability becomes complex, particularly in distributed AI architectures~\cite{das2025llm_security}. For instance, when AI systems are developed by multiple organizations or individuals across different regions, the absence of standardized regulations across jurisdictions can lead to inconsistent governance and legal ambiguities~\cite{kumar2024large}. Moreover, privacy and data protection remain critical concerns, as AI systems rely on vast amounts of data, increasing the risk of unauthorized access, misuse, or breaches~\cite{farahani2024artificial}. The rapid evolution of AI often outpaces legislative efforts, creating regulatory gaps that could be exploited. To mitigate these risks and enable the responsible large-scale deployment of AI systems, it is essential to establish clear legal frameworks, such as robust data protection laws, and continuously adapt regulatory policies in response to AI’s advancing capabilities.

\subsection{Environmental Sustainability} The rapid advancement of AI and its integration with wireless networks have raised significant environmental concerns, primarily due to the substantial energy consumption required for training and deploying large-scale AI models~\cite{ding2024sustainable}. For instance, training BigScience's 176-billion-parameter BLOOM model generated approximately 25 tons of carbon dioxide emissions, which is equivalent to driving a car around the Earth five times~\cite{luccioni2023estimating}. As AI models continue to expand in scale and complexity, their growing carbon footprint poses a serious challenge to environmental sustainability. To mitigate these impacts, it is essential to develop energy-efficient AI architectures, optimize computational processes, and incorporate renewable energy sources into AI training and inference.

\section{Conclusion}
\label{sec:conclusion}
This survey begins by reviewing recent advancements in both wireless networks and RL-based LLMs, then takes the first step in exploring how these two domains can mutually empower each other. We provide a comprehensive analysis of this mutual empowerment, including motivations, challenges, and potential solutions. In addition, we examine relevant applications, discuss future research directions, and evaluate the societal impact of this convergence. The primary objective of this survey is to offer readers innovative insights into the interplay between wireless networks and RL-based LLMs, distinguishing our work from existing surveys that primarily focus on LLM applications. Moving forward, we aim to further investigate this topic and provide a more comprehensive analysis.



\bibliographystyle{ACM-Reference-Format}
\bibliography{ref}

\appendix

\end{document}